\documentclass{sharpa}

\newcommand{\TODO}[1]{\textbf{\color{red}[TODO: #1]}}
\renewcommand{\TODO}[1]{}

\usepackage{newtxtt}
\definecolor{hanblue}{rgb}{0.27, 0.42, 0.81}
\hypersetup{
    pdftitle={SHAPE: Discrete Supervision and Parallel Continuous Streams for Learning Dexterous Manipulation from Egocentric Human Data},
    pdfauthor={Yaping Li, Zhihong Liu, Jiacheng Li, Yi Yang, Wenchao Yue, Mingzhe Ye, Xiangcheng Liu, Yiyang Chen, Haotian He, Qi Zhang, Haonan Wang, Dahua Lin, Zhijie Deng},
    colorlinks=true,
    linkcolor=hanblue,
    urlcolor=hanblue,
    citecolor=hanblue,
    anchorcolor=hanblue}
\AtBeginDocument{%
  \urlstyle{sf}%
}

\renewcommand{\cite}{\citep}

\titleformat*{\section}{\Large\sffamily\bfseries\color{accentcolor}\nobreak}
\titleformat*{\subsection}{\large\sffamily\bfseries\color{accentcolor}\nobreak}
\titleformat*{\subsubsection}{\normalsize\sffamily\bfseries\color{accentcolor}\nobreak}

\makeatletter
\@ifundefined{IfSocketExistsTF}{}{%
  \IfSocketExistsTF{hyp/raisedlink}{%
    \NewSocketPlug{hyp/raisedlink}{ttldefer}{\gdef\ttl@Hy@saveanchor{\Hy@raisedlink{#1}}}%
    \def\sharpa@ttl@refstepcounter#1{%
      \AssignSocketPlug{hyp/raisedlink}{ttldefer}%
      \refstepcounter{#1}%
      \AssignSocketPlug{hyp/raisedlink}{identity}}%
    \AtBeginDocument{\let\ttl@Hy@refstepcounter\sharpa@ttl@refstepcounter}%
  }{}}
\makeatother

\patchcmd{\mymaketitle}{\authorlist\par}{{\baselineskip=14pt\authorlist\par}}{}%
  {\ClassWarning{sharpa}{Could not set the leading of the author list}}

\newlength{\narrowtablewidth}
\usepackage{amsmath}
\usepackage{amssymb}
\usepackage{xspace}
\usepackage{float}
\usepackage{makecell}
\usepackage{pifont}
\usepackage{CJKutf8}

\newcommand{\cmark}{\ding{51}}
\newcommand{\xmark}{\ding{55}}

\newcommand{\Lfast}{\mathcal{L}_{\text{FAST}}}
\newcommand{\Lvideo}{\mathcal{L}_{\text{video}}}
\newcommand{\Ltext}{\mathcal{L}_{\text{text}}}

\newif\ifshowtodos
\newif\ifshowtbd
\showtodosfalse
\showtbdfalse

\definecolor{tbdcolor}{RGB}{204,102,0}
\definecolor{todocolor}{RGB}{200,30,30}

\DeclareRobustCommand{\method}{SHAPE\xspace}

\newsavebox{\todolinesbox}

\title{SHAPE: Discrete Supervision and Parallel Continuous\\Streams for Learning Dexterous Manipulation\\from Egocentric Human Data}

\author[1,2]{Yaping Li}
\author[1,3]{Zhihong Liu}
\author[1]{Jiacheng Li}
\author[1,3]{Yi Yang}
\author[1,2]{Wenchao Yue}
\author[1,3]{\linebreak Mingzhe Ye}
\author[1]{Xiangcheng Liu}
\author[1,3]{Yiyang Chen}
\author[1]{Haotian He}
\author[1]{\linebreak Qi Zhang}
\author[1]{Haonan Wang}
\author[2]{Dahua Lin}
\author[3]{Zhijie Deng}

\affiliation[1]{Sharpa Robotics}
\affiliation[2]{The Chinese University of Hong Kong}
\affiliation[3]{\mbox{SJTU-Sharpa Joint Lab for Embodied AI, SJTU}}

\abstract{Egocentric human recordings are far cheaper than dexterous robot demonstrations, promising stronger pretraining and direct skill learning without task-specific robot data. Yet humans and robots differ in morphology and kinematics, so human recordings lack the joint positions and actuator commands that supervise robot policies, providing only retargeted end-effector (EEF) actions. This raises two questions: how human data should supervise the policy, and how a dexterous robot should read out actions at deployment. We address both in \textbf{SHAPE} (\textbf{S}upervision from \textbf{H}uman \textbf{A}ction tokens with \textbf{P}arallel \textbf{E}xecutable streams), a vision-language-action (VLA) model whose backbone predicts discrete action tokens of EEF chunks as an auxiliary task, while a shared flow-matching expert outputs two executable streams, EEF and joint. On a bimanual robot with 22-DoF hands, a matched study shows that \textbf{discrete supervision helps learn from human data}, driving backbone gradients and shaping action-informative representations across domains. Under standard control pipelines at deployment, \textbf{direct joint readout excels at fine-grained control with robot demonstrations}, scoring 0.80 across six tasks against 0.63 for $\pi_{0.5}$, whereas \textbf{IK-based EEF readout excels at human-to-robot transfer}. Pretrained with human data and discrete tokens, \textbf{SHAPE learns new tasks from human demonstrations without task-specific robot data}. The project page is available \href{https://li-yaping.github.io/SHAPE/}{here}.
}

\renewcommand{\titlelogo}{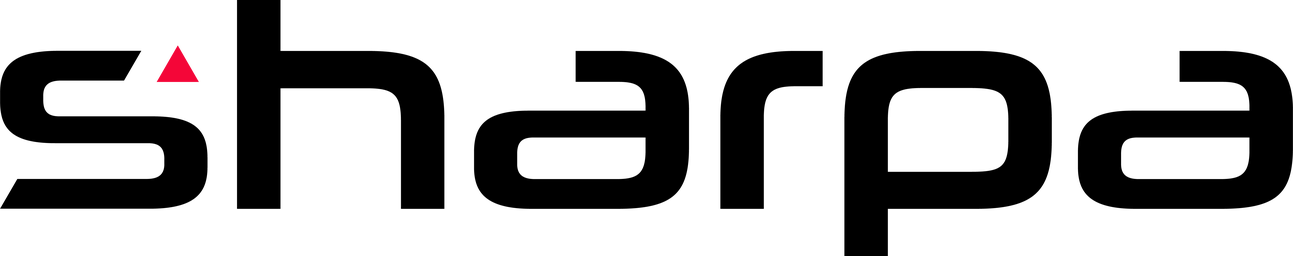}
\renewcommand{\titlelogowidth}{2.8cm}
\renewcommand{\titlelogoraise}{-16.7pt}

\begin{document}

\maketitle

\begin{figure}[!ht]
  \centering
  \includegraphics[width=\textwidth]{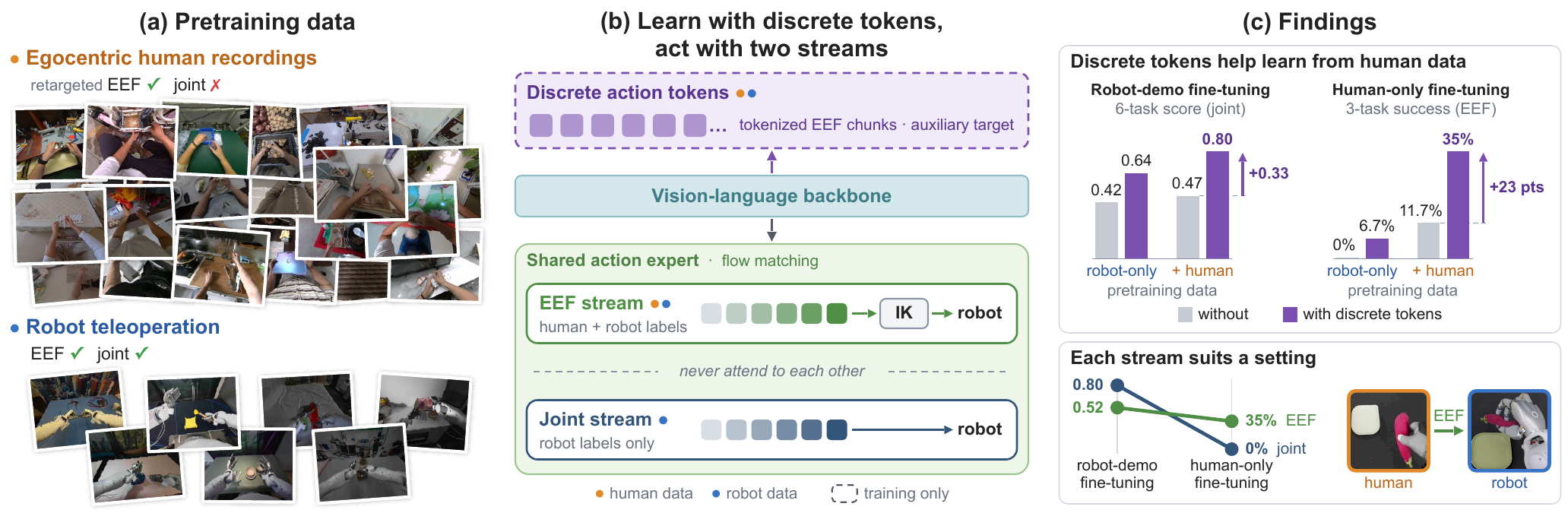}
  \caption{\textbf{SHAPE learns dexterous manipulation from egocentric human data.} (a) Pretraining pairs human recordings, providing only retargeted end-effector (EEF) actions, with robot teleoperation (EEF and joint actions). (b) The backbone predicts discrete EEF tokens across both domains as an auxiliary target, while a shared flow-matching expert outputs independent EEF and joint streams. (c) Adding human data and discrete tokens in pretraining improves policies under both robot-demo and human-only fine-tuning (top), where direct joint and IK-based EEF readouts respectively excel (bottom).}
  \label{fig:teaser}
\end{figure}

\section{Introduction}
\label{sec:intro}

Vision-language-action (VLA) models learn general manipulation skills by pretraining on large collections of robot demonstrations~\cite{rt2,openvla,pi0,gr00tn1,gr00tn17}. For dexterous bimanual manipulation such demonstrations are expensive: each one requires an operator to teleoperate two arms and two multi-fingered hands. Egocentric human recordings with tracked hand poses are much cheaper to collect at scale~\cite{hoi4d,holoassist,taco,hot3d,egodex,egoverse} and cover the same everyday activities. Such recordings hold two promises for dexterous manipulation: \textbf{improving pretraining} when downstream robot demonstrations are available, and \textbf{enabling a policy to learn new tasks directly from human demonstrations alone}, with zero task-specific robot data.

Fulfilling both promises on dexterous hands is challenging because humans and robots differ in morphology and kinematics. Human recordings lack the joint positions and actuator commands that supervise robot policies: tracked hands can be retargeted to a shared end-effector (EEF) space, whereas mapping them to robot joints via inverse kinematics (IK) relies on embodiment-specific correspondences and yields out-of-distribution arm postures. This action gap creates two dilemmas. First, \textbf{how should human data supervise the policy?} Many methods supervise shared EEF actions via a continuous action decoder~\cite{egomimic,egovla,hat,egozero,duet}, sending gradients to the backbone only through that decoder; recent VLAs also train the backbone on discrete action tokens~\cite{beingh0,pi05,ki}, including on human data~\cite{pi05ego}. Yet because they differ in data and architecture at once, or combine both losses without separating them~\cite{pi05ego}, whether and how discrete supervision helps learn from human data remains unclear. Second, \textbf{how should a dexterous robot read out actions at deployment?} Unlike 1-DoF grippers, multi-fingered hands introduce high redundancy between fingertip poses and finger joints. Prior policies add a robot-only joint output beside a shared EEF output but execute only the joint output~\cite{egomimic,duet,nexusvla}. Yet the two outputs excel in different settings: direct joint readout is more accurate when fine-tuned with robot demonstrations, while EEF readout via IK transfers better from human demonstrations.

In this work, we find that \textbf{auxiliary discrete action supervision helps the backbone learn from egocentric human data}, and we pair it with \textbf{a dual-stream action expert that keeps both EEF and joint streams executable}: the joint stream for precise control with robot demonstrations, and the EEF stream for human-to-robot transfer. As shown in Fig.~\ref{fig:teaser}, in \textbf{SHAPE} (\textbf{S}upervision from \textbf{H}uman \textbf{A}ction tokens with \textbf{P}arallel \textbf{E}xecutable streams), the backbone predicts discrete FAST tokens~\cite{fast} of each EEF chunk in one vocabulary for human and robot samples, supervising the backbone directly. In parallel, a shared flow-matching action expert independently outputs an EEF stream, supervised by both human and robot data, and a joint stream, supervised only by robot data in pretraining. Because both streams share the backbone and expert, human pretraining improves the joint stream without human joint labels, while either stream can drive the robot at test time.

To evaluate both choices under controlled conditions, we pretrain four models in a matched 2$\times$2 design that toggles human data and shared $\Lfast$ while fixing the architecture, robot data, and training budget. The human data comprise 2.8k hours of egocentric recordings, and the robot data 228 hours of teleoperation of two 7-DoF arms with 22-DoF hands. We evaluate the four models (reading out both streams) and two baselines in 1,680 real-robot trials, offline on held-out tasks, and via their representations and gradients. In simulation, we also evaluate all four models alongside both baselines on a second robot absent from pretraining, across both real-to-sim and cross-embodiment gaps.

Our contributions are:

\begin{itemize}
\item We find that \textbf{discrete action supervision helps a VLA learn dexterous manipulation from egocentric human data}, driving backbone gradients and task-aligned action representations across domains.
\item We propose \textbf{a dual-stream action expert with executable EEF and joint streams}, transferring human pretraining to joint control without human joint labels while supporting flexible stream selection at deployment.
\item Pretrained with human data and discrete tokens, SHAPE excels in both downstream regimes: fine-tuned on human demos alone, \textbf{it learns new tasks without task-specific robot data}; fine-tuned on robot demos, it scores 0.80 across six tasks against 0.63 for $\pi_{0.5}$.
\end{itemize}

\section{Related Work}
\label{sec:related}

\paragraph{Vision-language-action models.}
Recent VLAs couple a pretrained vision-language model (VLM) with an action decoder, either by emitting discretized action tokens~\cite{rt2,openvla,fast} or by attaching a continuous diffusion or flow-matching expert~\cite{pi0,pi05,gr00tn1,cogact,ctvam}, with extensions for tactile feedback or sensory prediction~\cite{tacforcing,star} and cross-embodiment pretraining~\cite{octo,rdt,hpt,beingh05}. $\pi_{0.5}$ with knowledge insulation~\cite{ki} also trains the VLM on FAST tokens while an expert produces continuous actions, but stops expert gradients from reaching the VLM and learns actions from robot data only. We keep both paths trainable and ask a different question: what discrete supervision does for learning from human data.

\paragraph{Learning manipulation from human videos.}
Human videos have been used to pretrain visual representations~\cite{ego4d,r3m,mvp,vc1} or action priors~\cite{videodex}, predict affordances, flow, point tracks, or keypoints~\cite{vrb,atm,track2act,kai,skelwam}, learn latent actions or visual foresight~\cite{lapa,gr00tn1,mantis,internvlaa1}, and, with tracked hands or motion-capture gloves, supervise actions in a shared EEF space~\cite{dexcap,egomimic,egovla,beingh0,beingh05,hrdt,phantom,dexwild,egozero,hat,pi05ego,psi0,emma,duet,motiontrans,egopi,egoscale,egoverse,exe,nexusvla}. Co-training does not always help: gains require diverse pretraining~\cite{pi05ego} or embodiment-specific networks~\cite{whatmatters}. While Phantom~\cite{phantom} renders a robot over the human arm and others align human and robot features~\cite{egobridge,innon}, $\pi_{0.5}$+ego~\cite{pi05ego} uses both FAST tokens and flow matching on human data for grippers without separating the two, and EgoLAP~\cite{egolap} supervises the VLM with structured language actions and motion rationales on bimanual grippers while stopping flow-expert gradients at the backbone~\cite{ki}. Using neither rendering, explicit feature alignment, nor gradient insulation, our matched 2$\times$2 pretraining study and continuous-head controls on a dexterous bimanual robot isolate what direct backbone supervision adds to learning from human data, when fine-tuning downstream tasks both with and without robot demonstrations.

\paragraph{Action representations for dexterous control.}
Action chunks can be tokenized~\cite{rt1,vqbet,fast} or modeled continuously with diffusion~\cite{diffusionpolicy} or flow matching~\cite{pi0}. For dexterous hands, policies act either in EEF and fingertip space, which transfers across embodiments via IK~\cite{dexcap}, or in joint space, which is precise but embodiment-specific~\cite{videodex,dexmv,anyteleop,dexpilot}; a large study on arms with grippers finds joint space more stable and task space better at generalizing~\cite{demystifying}. EgoMimic~\cite{egomimic,act}, DUET~\cite{duet}, and the concurrent Nexus-VLA~\cite{nexusvla} add a robot-only joint output beside a shared EEF output, but execute only the joint output. We keep both streams executable and, from the same checkpoints, find direct joint readout more accurate when fine-tuned with robot demonstrations, while wrist-frame EEF readout via IK transfers better from retargeted human demonstrations.

\section{Method}
\label{sec:method}

\subsection{Problem setup}
\label{sec:setup}

We learn a policy that maps an observation $o_t$ and an instruction $\ell$ to an action chunk $A_t = [a_t, \ldots, a_{t+H-1}]$ with $H = 64$, from two data sources. Robot teleoperation $\mathcal{D}_r$ provides head and wrist camera images, proprioception, and actions in both EEF and joint space. Egocentric human recordings $\mathcal{D}_h$ provide head-camera images and tracked 3D hand poses, but lack executable joint states and actuator commands; we retarget the tracked hands to derive EEF actions.

\paragraph{Action parameterization.}
Both action spaces are relative to the current state. For each hand, let $T_t \in \mathrm{SE}(3)$ be the wrist pose at time $t$ and $p_{t}$ the positions of the five fingertips in the wrist frame, and let $q_{t}$ be the joint positions. For $k = 0, \ldots, H-1$, the EEF action contains the wrist pose expressed in the current wrist frame, $T_t^{-1} T_{t+k}$, and the fingertip offsets $p_{t+k} - p_t$; the joint action is the offset $q_{t+k} - q_t$. In pretraining, each data source is normalized with its own statistics. For human recordings, the hand is first estimated with the MANO hand model~\cite{mano} and then retargeted to the robot hand, and the fingertip positions are those of the retargeted hand. Expressing EEF actions in the current wrist frame cancels camera extrinsics and decouples the action target from the underlying arm joint configuration.

\begin{figure}[t]
  \centering
  \includegraphics[width=\textwidth]{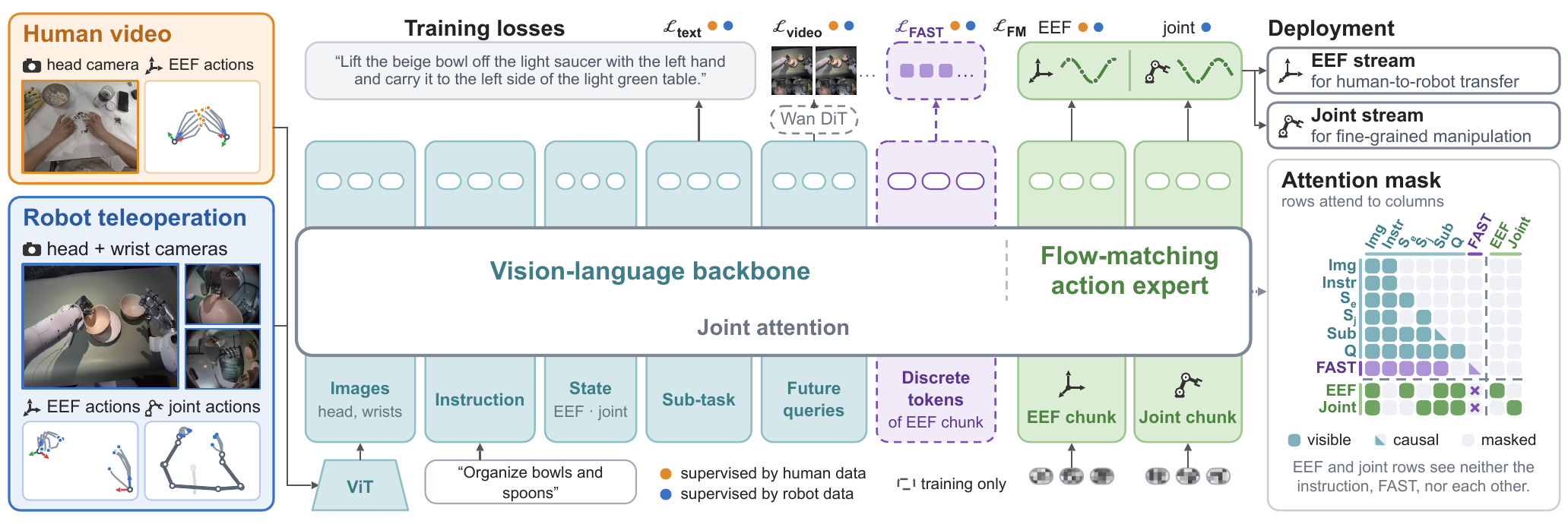}
  \caption{\textbf{Overview of SHAPE.} Human recordings (orange) provide head images and EEF actions; robot data (blue) add wrist images and joint actions. The backbone predicts discrete FAST tokens of the EEF chunk ($\Lfast$) as an auxiliary task alongside the sub-task ($\Ltext$) and future frames ($\Lvideo$). A shared flow-matching expert reads the backbone at every layer and independently denoises an EEF and a joint chunk, isolated by the attention mask. At deployment, training-only heads (dashed) are omitted, and either stream can drive the robot: the joint stream with robot demos, and the EEF stream for human-to-robot transfer.}
  \label{fig:method}
\end{figure}

\subsection{Backbone and dual-stream action expert}
\label{sec:arch}

As shown in Fig.~\ref{fig:method}, a single backbone and a single action expert serve both embodiments; only the output streams differ. The backbone is Qwen3-VL-4B~\cite{qwen3vl}. Its input contains image tokens, instruction tokens, and a state prefix of two tokens that embed the current EEF state and, for robot samples, the joint state. During training each state token is dropped independently with probability 0.5, so that the policy can run with or without it. Following recent VLAs, the backbone also predicts the current sub-task as text ($\Ltext$; \citealp{pi05}) and processes 64 learnable future queries that condition a Wan video diffusion head~\cite{wan} to predict frames about 1\,s ahead ($\Lvideo$; \citealp{mantis,wla,internvlaa15}). The action expert is a 0.6B transformer that attends to the backbone keys and values at every layer~\cite{pi0}. Appendix~\ref{supp:impl} gives architectural details.

The action expert is shared by two parallel output streams. The 55-d EEF stream represents each side by its 3D wrist position, a 6-d wrist rotation, and five 3D fingertip positions, and adds 7 torso and neck joints. The 65-d joint stream covers two 7-DoF arms, two 22-DoF hands, and the same 7 torso and neck joints. Each stream has its own input and output projections and is trained with a flow-matching loss~\cite{flowmatching}, $\mathcal{L}^{\text{EEF}}_{\text{FM}}$ or $\mathcal{L}^{\text{joint}}_{\text{FM}}$. In the attention mask (Fig.~\ref{fig:method}, bottom right), each stream reads the image, sub-task, and query tokens and its own state token, but neither the instruction nor the other stream, so the two streams can be executed independently at deployment. A per-sample mask removes losses whose labels are missing: in pretraining, human samples supervise the EEF stream but not the joint stream, yet still shape the joint stream through the shared backbone and expert.

\subsection{Discrete action supervision}
\label{sec:discrete}

With the continuous streams alone, action supervision reaches the backbone only through the action expert, and human samples provide no joint labels in pretraining. To supervise the backbone directly through its native token head on every sample, we predict discrete FAST tokens~\cite{fast,pi05,ki,pi05ego} of each normalized EEF chunk as an auxiliary task. Avoiding dense per-step binning~\cite{rt1} or learned vector quantizers~\cite{vqbet}, FAST compresses chunks via a discrete cosine transform and byte-pair encoding, with its 1,024-token vocabulary mapped onto the backbone's. Sharing one vocabulary across human and robot chunks lets the backbone predict them autoregressively under $\Lfast$ (compared against continuous backbone heads in Sec.~\ref{sec:contihead}). In the attention mask (Fig.~\ref{fig:method}), the action expert cannot read the FAST tokens, preventing shortcut copying so they can be omitted at deployment.

\subsection{Training}
\label{sec:training}

The primary training objectives are optimized jointly:
\begin{equation}
  \mathcal{L} = \mathcal{L}^{\text{EEF}}_{\text{FM}} + m_J\, \mathcal{L}^{\text{joint}}_{\text{FM}}
  + \lambda_F\, \Lfast + \lambda_V\, \Lvideo + \lambda_T\, \Ltext,
  \label{eq:loss}
\end{equation}
where $m_J$ masks samples without joint labels, $\lambda_F = 0.1$, $\lambda_V = 1$, and $\lambda_T = 0.2$ (full hyperparameters in Appendix~\ref{supp:impl}). Pretraining samples human and robot data 1\,:\,1; robot-demo fine-tuning retains each model's pretraining losses (Secs.~\ref{sec:benchmark}, \ref{sec:sim}), whereas human-only fine-tuning disables $\Lfast$ across all models to isolate pretraining and compare both streams equally (Sec.~\ref{sec:h2r}).

\subsection{Deployment}
\label{sec:deploy}

At test time, only the backbone and the action expert run; the auxiliary FAST tokens are omitted (see Appendix~\ref{supp:fastdec} for an analysis of token decoding), and each rollout executes one of the two continuous streams (Fig.~\ref{fig:method}, top right). The joint stream commands the joint controllers directly, whereas the EEF stream converts predicted wrist poses and fingertip positions into joint targets via inverse kinematics. Keeping both streams executable within a single model allows flexible stream selection at deployment and lets us compare direct joint readout against IK-based EEF readout from the same checkpoint across both deployment regimes (Sec.~\ref{sec:stream}).

\section{Experimental Setup}
\label{sec:exp}

\paragraph{Platform and data.}
We use a dual-arm robot with two 7-DoF arms and two 22-DoF dexterous hands, a head camera, and two wrist cameras. Pretraining uses 2.8k hours of egocentric human recordings with tracked hands and 228 hours of robot teleoperation (320 and 105 tasks, respectively); Table~\ref{tab:models} summarizes both corpora and defines the four matched models.

\begin{table}[ht]
  \caption{\textbf{Pretraining data and the four matched pretrained models.} Left: pretraining corpora and action supervision targets. Right: the four matched models, which share the architecture, robot data, and training budget.}
  \label{tab:models}
  \centering\footnotesize
  \begin{minipage}[t]{0.55\textwidth}
    \setlength{\tabcolsep}{3.5pt}
    \begin{tabular*}{\linewidth}{@{\extracolsep{\fill}}lcccccc@{}}
      \toprule
      Source & Tasks & Episodes & Hours & EEF & Joint & FAST \\
      \midrule
      Human egocentric & 320 & 73.1k & 2,811 & \cmark & -- & \cmark \\
      Robot teleoperation & 105 & 10.5k & 228 & \cmark & \cmark & \cmark \\
      \bottomrule
    \end{tabular*}
  \end{minipage}\hfill
  \begin{minipage}[t]{0.42\textwidth}
    \setlength{\tabcolsep}{4pt}
    \begin{tabular*}{\linewidth}{@{\extracolsep{\fill}}lccl@{}}
      \toprule
      Model & Human data & $\Lfast$ & Used as \\
      \midrule
      SHAPE & \cmark & \cmark & full model \\
      w/o FAST & \cmark & \xmark & ablation \\
      w/o human & \xmark & \cmark & ablation \\
      w/o both & \xmark & \xmark & ablation \\
      \bottomrule
    \end{tabular*}
  \end{minipage}
\end{table}

\begin{figure}[t]
  \centering
  \includegraphics[width=\textwidth]{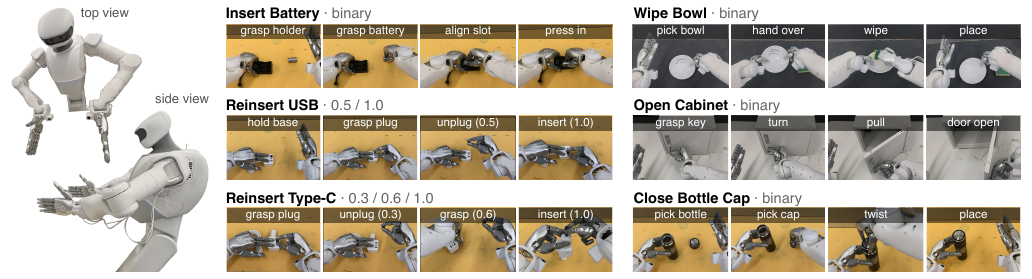}
  \caption{\textbf{Real-robot tasks with robot demonstrations.} The left panel shows top and side views of the robot, while the right panels illustrate the four stages of each task, with partial credit indicated in parentheses for Reinsert USB and Reinsert Type-C.}
  \label{fig:tasks}
\end{figure}

\paragraph{Tasks.}
Six real-robot tasks have robot demonstrations (Fig.~\ref{fig:tasks}): precise insertion (Insert Battery), unplugging and reinserting a connector (Reinsert USB, Reinsert Type-C), a handover (Wipe Bowl), an articulated object (Open Cabinet), and in-hand twisting (Close Bottle Cap), each fine-tuned on 196--447 teleoperated demonstrations. Three human-to-robot (H2R) tasks, Pick Eggplant, Pick Cup, and Stack Cups, have zero task-specific robot demonstrations and are fine-tuned on 188--1,187 egocentric human demonstrations recorded with a head-mounted device.

\paragraph{Protocol.}
Each policy is evaluated for 20 trials per task with randomized object placements. Multi-stage USB and Type-C receive stage credit (USB: 0.5 unplug, 1.0 insert; Type-C: 0.3 unplug, 0.6 grasp, 1.0 insert; stage counts in Appendix~\ref{supp:real}); all other tasks are binary. With robot demonstrations, we compare six-task scores. We report 95\% confidence intervals (CIs) and two-sided tests throughout (Appendix~\ref{supp:stats}).

\paragraph{Baselines.}
We compare with $\pi_{0.5}$~\cite{pi05} and GR00T N1.7~\cite{gr00tn1,gr00tn17}, which is also pretrained on 20k hours of egocentric human video~\cite{egoscale}. Both start from their public checkpoints and are fine-tuned on the same demonstrations in the same joint action spaces, for 30 epochs on the real robot and 100k steps (batch 256) in simulation. More baseline and fine-tuning details are given in Appendix~\ref{supp:baselines}.

\section{Closed-Loop Results}
\label{sec:results}

\subsection{Tasks with robot demonstrations}
\label{sec:benchmark}

\begin{table}[t]
  \caption{\textbf{Real-robot results with robot demonstrations.} Each task is evaluated over 20 trials; w/o both omits both human data and $\Lfast$ in pretraining. Cells give the number of successes, or the mean partial credit for USB and Type-C, and Score is the six-task mean. The chart on the right shows the scores with 95\% intervals.}
  \label{tab:benchmark}
  \centering\footnotesize
  \begin{minipage}[c]{\dimexpr\textwidth-2in-8pt\relax}
    \setlength{\tabcolsep}{2pt}
    \begin{tabular*}{\linewidth}{@{\extracolsep{\fill}}lcccccccc@{}}
      \toprule
      \multirow{2}{*}{Method} & \multirow{2}{*}{Stream} & Insert & Reinsert & Reinsert & Wipe & Open & Close & \multirow{2}{*}{Score~$\uparrow$} \\
      & & Battery & USB & Type-C & Bowl & Cabinet & Bottle Cap & \\
      \midrule
      GR00T N1.7 & Joint & 5/20 & 0.18 & 0.52 & 12/20 & \textbf{18/20} & 12/20 & 0.51 \\
      $\pi_{0.5}$ & Joint & 13/20 & 0.25 & 0.62 & 13/20 & 17/20 & 15/20 & 0.63 \\
      \midrule
      \multirow{2}{*}{SHAPE} & EEF & 6/20 & 0.15 & 0.54 & 14/20 & 17/20 & 12/20 & 0.52 \\
      & Joint & \textbf{16/20} & \textbf{0.55} & \textbf{0.64} & \textbf{18/20} & \textbf{18/20} & \textbf{20/20} & \textbf{0.80} \\
      \midrule
      \multirow{2}{*}{w/o human} & EEF & 7/20 & 0.15 & 0.42 & 10/20 & 15/20 & 13/20 & 0.47 \\
      & Joint & 9/20 & 0.25 & 0.55 & 16/20 & \textbf{18/20} & 18/20 & 0.64 \\
      \midrule
      \multirow{2}{*}{w/o FAST} & EEF & 6/20 & 0.00 & 0.20 & 1/20 & 13/20 & 2/20 & 0.22 \\
      & Joint & 11/20 & 0.08 & 0.51 & 9/20 & 11/20 & 14/20 & 0.47 \\
      \midrule
      \multirow{2}{*}{w/o both} & EEF & 3/20 & 0.03 & 0.11 & 5/20 & 11/20 & 2/20 & 0.20 \\
      & Joint & 6/20 & 0.33 & 0.35 & 10/20 & 12/20 & 9/20 & 0.42 \\
      \bottomrule
    \end{tabular*}
  \end{minipage}\hfill
  \begin{minipage}[c]{2in}
    \raisebox{-6pt}[\height][0pt]{\includegraphics[width=\linewidth]{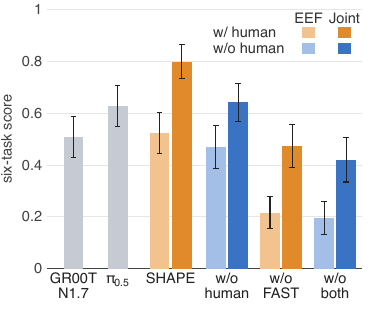}}
  \end{minipage}
\end{table}

\paragraph{Comparison with VLA baselines.}
With the joint stream, SHAPE scores 0.80 over the six tasks, against 0.63 for $\pi_{0.5}$ and 0.51 for GR00T N1.7 (Table~\ref{tab:benchmark}). It is best or tied on every task, with the largest margins over $\pi_{0.5}$ on Reinsert USB, Wipe Bowl, and Close Bottle Cap, and over GR00T N1.7 on Insert Battery; Open Cabinet is near ceiling. Counting strict full-task completions (2/20 final reinsertions on both connectors; Appendix~\ref{supp:real}), SHAPE still leads at 76/120 (63.3\%) vs.\ 61/120 (50.8\%) for $\pi_{0.5}$ and 51/120 (42.5\%) for GR00T N1.7, though millimeter-tolerance connector reinsertion remains challenging.

\paragraph{Where does the gain come from?}
The matched ablations show that discrete supervision and human pretraining contribute complementary gains rather than architectural advantages alone. Without human data, our model matches $\pi_{0.5}$ (0.64 vs.\ 0.63). Including $\Lfast$ across pretraining and fine-tuning raises the six-task score by 0.22--0.33 across both streams, with or without human data ($p < 0.001$; Secs.~\ref{sec:h2r} and \ref{sec:analysis} isolate pretraining $\Lfast$ directly). Holding $\Lfast$ fixed, adding human pretraining data raises the joint-stream score from 0.64 to 0.80 (+0.16, $p = 0.002$), improving all five non-ceiling tasks and lifting success on the four binary tasks from 76.3\% (61/80) to 90.0\% (72/80); together, both ingredients raise the joint-stream score by +0.38 over w/o both (0.42 to 0.80).

\begin{figure}[t]
  \centering
  \includegraphics[width=\textwidth]{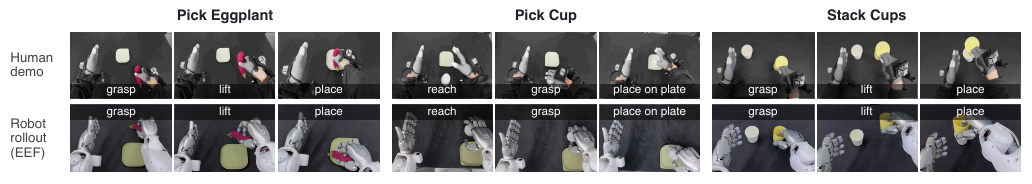}
  \caption{\textbf{Human-to-robot transfer with zero task-specific robot demonstrations.} Top: egocentric human demonstrations, the only fine-tuning data for each task. Bottom: robot rollouts of SHAPE with the EEF stream.}
  \label{fig:h2r}
\end{figure}

\subsection{Human-to-robot transfer}
\label{sec:h2r}

\paragraph{Setup.}
For each H2R task we fine-tune every pretrained model on egocentric human demonstrations only, masking out wrist camera inputs as detailed in Appendix~\ref{supp:h2r-setup}, and deploy it on the robot with zero task-specific robot demonstrations (Fig.~\ref{fig:h2r}). Fine-tuning supervises the EEF stream with human EEF actions and the joint stream with retargeted arm and hand joints, and both streams are read out from the same fine-tuned checkpoint. Because FAST tokens encode only EEF actions, $\Lfast$ is disabled for all four models during fine-tuning, comparing both streams on equal terms and isolating the effect of pretraining.

\begin{table}[t]
  \caption{\textbf{Human-to-robot transfer results with zero task-specific robot demonstrations.} Successes out of 20 trials per task on Pick Eggplant, Pick Cup, and Stack Cups; all rows use the EEF stream except the indented joint-stream row of SHAPE. The chart on the right shows the total success rates with 95\% intervals.}
  \label{tab:h2r}
  \centering\footnotesize
  \begin{minipage}[c]{\narrowtablewidth}
    \setlength{\tabcolsep}{2.2pt}
    \begin{tabular*}{\linewidth}{@{\extracolsep{\fill}}lcccccc@{}}
      \toprule
      Pretraining & Hum. & FAST & Egg. & Cup & Stack & \makecell[t]{Total\\(success rate)} \\
      \midrule
      w/o both & \xmark & \xmark & 0 & 0 & 0 & 0/60 (0\%) \\
      w/o human & \xmark & \cmark & 0 & 3 & 1 & 4/60 (6.7\%) \\
      w/o FAST & \cmark & \xmark & 3 & 2 & 2 & 7/60 (11.7\%) \\
      SHAPE & \cmark & \cmark & \textbf{10} & \textbf{8} & \textbf{3} & \textbf{21/60 (35\%)} \\
      \quad joint stream & \cmark & \cmark & 0 & 0 & 0 & 0/60 (0\%) \\
      \bottomrule
    \end{tabular*}
  \end{minipage}\hfill
  \begin{minipage}[c]{\dimexpr\textwidth-\narrowtablewidth-12pt\relax}
    \includegraphics[width=\linewidth]{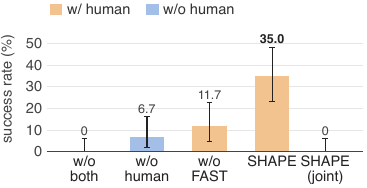}
  \end{minipage}
\end{table}

\paragraph{Combining human data and discrete supervision performs substantially best.}
Table~\ref{tab:h2r} summarizes the H2R transfer results. Without downstream robot demonstrations, either ingredient alone achieves limited transfer: robot-only pretraining reaches 0/60 (0\%) without $\Lfast$ and 4/60 (6.7\%) with it, while human pretraining without $\Lfast$ reaches 7/60 (11.7\%); combining both in SHAPE reaches 21/60 (35.0\%), a +28.3-point gain from human data ($p < 0.001$).

\paragraph{Pretraining $\Lfast$ drives transfer.}
On the human-pretrained models, pretraining with $\Lfast$ raises transfer success from 7/60 (11.7\%) to 21/60 (35.0\%, +23.3 points, $p = 0.004$), improving every task. Because all four models share identical fine-tuning supervision with $\Lfast$ disabled, this +23.3-point gain is consistent with a pretraining effect of $\Lfast$ (analyzed in Sec.~\ref{sec:repr}).

\paragraph{The EEF stream adapts to object placement.}
With the EEF stream, human-pretrained models adapt to varied object placements: SHAPE reaches toward the object's position and places it at the target across different layouts rather than replaying a fixed trajectory. Most failures occur at the grasp, where the hand reaches the object but closes too loosely to lift it, reflecting morphological and kinematic gaps that retargeting does not fully bridge. The joint stream instead replays the demonstrated motion, barely adapts to object position, and never succeeds (Sec.~\ref{sec:stream}).

\subsection{Which stream to deploy?}
\label{sec:stream}

\paragraph{Shared training, flexible readout.}
Human recordings never supervise the joint stream in pretraining, reaching it only through the shared backbone and action expert (Sec.~\ref{sec:arch}). Even so, human pretraining with $\Lfast$ raises the joint stream's six-task score from 0.64 to 0.80 on the real robot (Sec.~\ref{sec:benchmark}) and its success rate by +7.0 points in simulation (Sec.~\ref{sec:sim}). Keeping both streams executable then lets us compare direct joint readout against IK-based EEF readout from the same fine-tuned checkpoint.

\paragraph{With robot demonstrations, direct joint readout excels at fine-grained control.}
From the same checkpoints, direct joint readout outperforms IK-based EEF readout in 23 of 24 model--task pairs (Table~\ref{tab:benchmark}), raising each model's six-task score by 0.17--0.28. Gains are largest on contact-rich tasks such as Close Bottle Cap, Wipe Bowl, and Insert Battery. In open-loop evaluation on the 52 validation episodes of the six tasks (Appendix~\ref{supp:data}), however, both streams---converting joint predictions via forward kinematics---achieve the same mean 3D Euclidean error over 64-step chunks (right wrist: 11.6 vs.\ 11.5\,mm; fingertips: 4.8 vs.\ 4.9\,mm). The gap instead arises during execution, consistent with under-constrained fingertip-to-joint IK: five 3D fingertip targets (15 values) under-constrain a 22-DoF hand, so small target or rotation errors can perturb the interphalangeal joints that govern grasp curl and contact. Direct joint readout commands all arm and finger joints without this inversion step.

\paragraph{With only human demonstrations, wrist-frame EEF readout enables transfer.}
Under our human-retargeting pipeline, the advantage reverses: across the four models and three H2R tasks, 0 of 240 joint-stream rollouts succeed vs.\ 32 of 240 for the EEF stream (Table~\ref{tab:h2r}; Appendix~\ref{supp:real}). In joint space, the 3D hand motion caused by an offset $\Delta q$ depends non-linearly on the arm posture $q_t$ via its Jacobian~\cite{demystifying}. Because per-frame retargeted human arm postures lie outside the robot's posture distribution, as analyzed in Appendix~\ref{supp:joint}, joint offsets learned around them fail from the robot's home posture. Even without a policy, replaying a demonstration's $\Delta q$ open loop from this home posture leaves the right hand 10.3\,cm from its EEF grasp target on average, stopping in front of and above the object. Wrist-frame EEF readout instead predicts 3D hand motion relative to the wrist and delegates joint resolution to IK at the robot's own posture, letting one dual-stream model serve both settings.

\subsection{A second robot in simulation}
\label{sec:sim}

Do human data and discrete supervision also help in simulation, on a robot absent from pretraining? We fine-tune the four matched models alongside $\pi_{0.5}$ and GR00T N1.7 on the 20.8k teleoperated demonstrations of the 24 GR-1 tabletop tasks of RoboCasa~\cite{robocasa,gr00tn1} under the same training budget and evaluate 50 episodes per task across three checkpoints (Table~\ref{tab:sim}), with per-task results reported in Appendix~\ref{supp:sim}. This benchmark tests transfer across both a real-to-sim gap, as all pretraining data are real-world recordings, and a cross-embodiment gap: the GR-1 humanoid has only a head camera and a 29-DoF action space comprising two 7-DoF arms, two 6-DoF Fourier hands (instead of 22-DoF hands), and a 3-DoF waist. SHAPE reaches 72.2\% overall, exceeding $\pi_{0.5}$ (64.8\%, +7.4 points, $p < 0.001$) and GR00T N1.7 (62.6\%, +9.6 points), whereas w/o human matches $\pi_{0.5}$ (65.2\% vs.\ 64.8\%) just as on the real robot. Across the 2$\times$2 models, the two ingredients act nearly additively: human pretraining adds +7.0 points with $\Lfast$ (65.2\% to 72.2\%, $p < 0.001$) and +7.6 points without it (62.0\% to 69.6\%), while $\Lfast$ adds +2.6 points with human data and +3.2 points without it, for a combined +10.2-point gain over w/o both.

\begin{table}[t]
  \caption{\textbf{A second robot in simulation.} Success rate (\%) on the 24 GR-1 tabletop tasks of RoboCasa, joint stream. Close and PnP are the 6 articulated and 18 pick-and-place tasks. The chart on the right shows the overall success rates with 95\% intervals.}
  \label{tab:sim}
  \centering\footnotesize
  \begin{minipage}[c]{\narrowtablewidth}
    \begin{tabular*}{\linewidth}{@{\extracolsep{\fill}}lccccc@{}}
      \toprule
      Method & Hum. & FAST & Close & PnP & All \\
      \midrule
      GR00T N1.7 & -- & -- & \textbf{69.2} & 60.4 & 62.6 \\
      $\pi_{0.5}$ & -- & -- & 51.4 & 69.2 & 64.8 \\
      \midrule
      w/o both & \xmark & \xmark & 61.4 & 62.2 & 62.0 \\
      w/o human & \xmark & \cmark & 62.6 & 66.1 & 65.2 \\
      w/o FAST & \cmark & \xmark & 57.1 & 73.8 & 69.6 \\
      SHAPE & \cmark & \cmark & 66.6 & \textbf{74.1} & \textbf{72.2} \\
      \bottomrule
    \end{tabular*}
  \end{minipage}\hfill
  \begin{minipage}[c]{\dimexpr\textwidth-\narrowtablewidth-12pt\relax}
    \includegraphics[width=\linewidth]{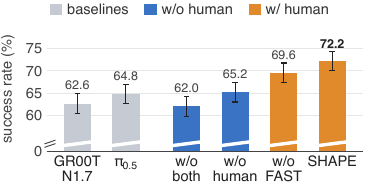}
  \end{minipage}
\end{table}
\section{Analysis}
\label{sec:analysis}

\begin{figure}[t]
  \centering
  \includegraphics[width=\textwidth]{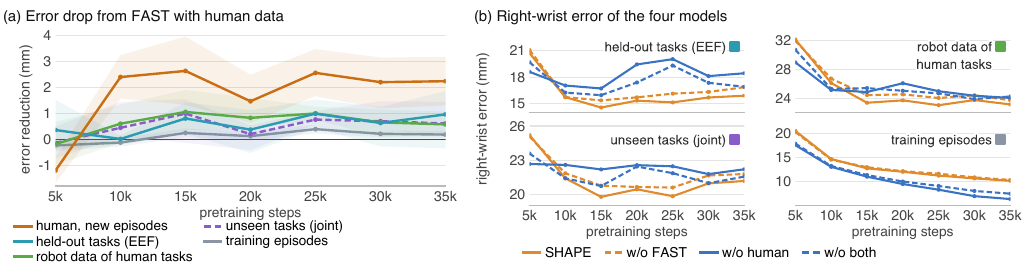}
  \caption{\textbf{Offline action error across pretraining steps.} (a) Right-wrist error reduction from adding $\Lfast$ with human data (w/o FAST minus SHAPE) across splits; bands show 95\% CIs. (b) Right-wrist errors of the four matched models (orange: with human data; blue: robot-only; dashed: without FAST; corner squares match split colors in (a)).}
  \label{fig:interaction}
\end{figure}

\subsection{Pretraining generalization across domains and streams}
\label{sec:offline}

To isolate pretraining from downstream fine-tuning recipes, we evaluate the four matched models every 5k pretraining steps on open-loop 3D position error over 64-step windows (Fig.~\ref{fig:interaction}, with all splits and body parts in Appendix~\ref{supp:offline}). On the narrower robot-only corpus (228\,h), adding $\Lfast$ lowers error on training episodes at every checkpoint yet raises zero-shot error on held-out robot tasks (+1.1\,mm on average), indicating overfitting under limited robot diversity. Paired with diverse human recordings (2.8k\,h), however, $\Lfast$ improves generalization throughout 10k--35k steps (Fig.~\ref{fig:interaction}a): it reduces right-wrist error by 1.5--2.6\,mm on unseen human episodes (all 95\% CIs $> 0$) and lowers error at every checkpoint on held-out robot tasks (EEF), robot data of human tasks, and unseen robot tasks (joint), giving SHAPE the lowest (or tied-lowest) error across all three held-out robot splits from 15k steps onward (Fig.~\ref{fig:interaction}b). Because the unseen-task gain holds when read out from the joint stream via forward kinematics, cross-stream transfer from human EEF data to joint control already emerges during pretraining.

\begin{figure}[t]
  \centering
  \includegraphics[width=\textwidth]{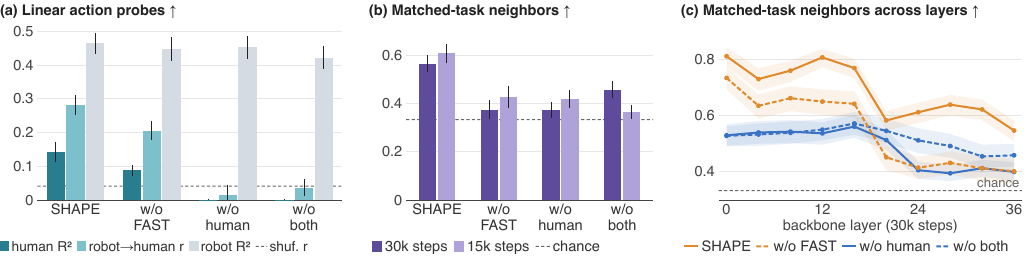}
  \caption{\textbf{Discrete supervision makes human frames more action-informative and closer to matched-task robot frames.} Shaded bands and error bars show 95\% confidence intervals. (a) Linear probe $R^2$ per domain and robot-to-human transfer $r$ on last-layer prefix-mean features at 30k steps (shuffled-label baseline dashed). (b) Share of a human frame's 10 nearest robot frames from its matched task across three task families (last-layer prefix mean, 15k and 30k steps). (c) Matched-task robot neighbor share across backbone layers at 30k steps on head-camera image tokens (layer 0: vision tower; layers 4--36: language model).}
  \label{fig:repr}
\end{figure}

\subsection{Action encoding and cross-embodiment task structure}
\label{sec:repr}

We next probe backbone features along two dimensions: linear action decodability and cross-embodiment task structure (Fig.~\ref{fig:repr}). On last-layer prefix-mean features (Fig.~\ref{fig:repr}a), models pretrained without human data carry negligible action signal on human frames ($R^2 \le 0.002$, transfer $r \le 0.05$). With human data, adding $\Lfast$ lifts the cross-validated $R^2$ of a linear action probe on human frames from 0.089 to 0.143 and raises the zero-shot correlation $r$ of a robot-trained probe on human frames from 0.205 to 0.281---an effect already clear at 15k steps that grows through the second half of the language model and reaches the joint stream of the expert, which human data never supervise in pretraining (Appendix~\ref{supp:repr}).

Beyond encoding motion magnitudes, $\Lfast$ organizes human and robot frames around shared task structure without explicit alignment loss (Fig.~\ref{fig:repr}b, c). Across three task families recorded in both domains---evaluated under one shared task-agnostic prompt with per-domain means removed---57\% of a human frame's ten nearest robot frames at the last layer come from its matched task for SHAPE, vs.\ 37\% for w/o FAST (+19 points, 95\% CI [+15, +23]), 37--46\% across all ablations, and 33\% by chance (Fig.~\ref{fig:repr}b). On head-camera image tokens from the vision tower through the language model (Fig.~\ref{fig:repr}c; Appendix~\ref{supp:matched}), SHAPE exceeds w/o FAST by 8--21 points at every probed layer, staying at 55--64\% in the deep layers vs.\ 40--45\% without $\Lfast$.

\begin{figure}[t]
  \centering
  \includegraphics[width=\textwidth]{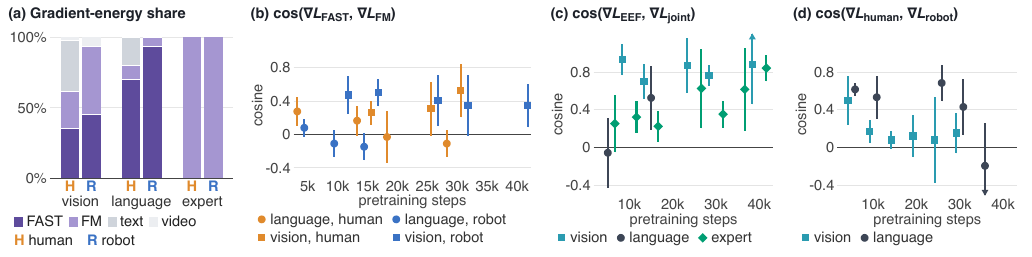}
  \caption{\textbf{How losses and data sources share the backbone.} (a) Share of each module's gradient energy (squared norm of the loss-weighted mean gradient across batches) at 30k steps on human and robot batches. (b--d) Cosines between mean gradients across pretraining steps (robot batches in (c)). Error bars show $\pm$2 standard errors where both mean gradients exceed batch-to-batch noise.}
  \label{fig:grad}
\end{figure}

\subsection{How losses, streams, and domains share gradients}
\label{sec:grad}

Measuring each loss's expected gradient energy across modules (Fig.~\ref{fig:grad}a; Appendix~\ref{supp:grad}) explains why $\Lfast$ reshapes the backbone even though $\mathcal{L}_{\text{FM}}$ already supervises EEF actions: $\mathcal{L}_{\text{FM}}$ supplies 100\% of the action expert's gradient energy, whereas $\Lfast$ dominates the language model---contributing 70\% and 93\% on human and robot batches at 30k steps ($\ge$47\% throughout) and 17--71\% on the vision tower. Where resolved above batch noise, $\nabla\Lfast$ and $\nabla\mathcal{L}_{\text{FM}}$ do not conflict (cosines $-0.15$ to +0.28 on the language model, +0.26 to +0.53 on the vision tower; Fig.~\ref{fig:grad}b), permitting joint training without stopping expert gradients.

Shared gradients likewise align across streams and domains. Where resolved, the cosine between EEF and joint flow-matching gradients is +0.70 to +0.93 on the vision tower and rises from +0.25 at 5k steps to +0.84 at 40k on the shared expert (Fig.~\ref{fig:grad}c, holding without FAST as well; Appendix~\ref{supp:grad}). Between human and robot batches, the cosine is positive at all six resolved checkpoints on the vision tower (+0.08 to +0.50) and at four of five on the language model (+0.43 to +0.69, with $-0.19 \pm 0.23$ at 35k; Fig.~\ref{fig:grad}d), letting human EEF updates reinforce features used by joint control.

\subsection{Gradient scale vs.\ discrete action tokens}
\label{sec:contihead}

Does any strong continuous action target on the backbone suffice? Replacing FAST tokens with 64 learnable queries in the same backbone slot---reading the identical prefix, mutually visible across the 64 steps, and masked from the action expert---to regress normalized EEF chunks via an L1 MLP head matched in initial backbone gradient norm ($\lambda = 0.1$) fails to sustain supervision: its backbone gradient norm shrinks to $0.11\text{--}0.13\times$ $\Lfast$ from 5k steps on, leaving layer-24 probes at the w/o FAST level (Appendix~\ref{supp:contihead}). Sustained direct backbone gradients are therefore necessary.

\begin{table}[t]
  \caption{\textbf{Gradient-matched continuous head vs.\ SHAPE.} Pretraining metrics at 30k steps (Appendix~\ref{supp:contihead}); NN is the matched-task neighbor share at layer 24 (L24; Fig.~\ref{fig:repr}c) and at the last layer (Fig.~\ref{fig:repr}b). H2R disables auxiliary action heads for all models (SHAPE vs.\ Cont.: $p = 0.055$); GR-1 averages 80k--100k ($p = 0.001$ w/~aux, $p = 0.11$ w/o~aux vs.\ SHAPE). The chart on the right contrasts last-layer probe $R^2$, matched-task neighbors (dashed: chance), and real-robot H2R transfer with 95\% intervals.}
  \label{tab:contihead}
  \centering\footnotesize
  \begin{minipage}[c]{0.63\textwidth}
    \setlength{\tabcolsep}{2.2pt}
    \begin{tabular*}{\linewidth}{@{\extracolsep{\fill}}lccc@{}}
      \toprule
      Metric & w/o FAST & Cont.\ head & SHAPE \\
      \midrule
      Backbone grad.\ norm / $\Lfast$ & -- & $0.98\times$ & $1.00\times$ \\
      Human new-ep.\ error (mm)~$\downarrow$ & 62.9 & \textbf{59.2} & 60.7 \\
      Held-out robot error (mm)~$\downarrow$ & 16.3 & 16.1 & \textbf{15.7} \\
      \midrule
      Human probe $R^2$ / transfer $r$~$\uparrow$ & 0.089 / 0.205 & \textbf{0.230} / \textbf{0.437} & 0.143 / 0.281 \\
      Matched-task NN, L24 / last (\%)~$\uparrow$ & 42 / 37 & \textbf{66} / 40 & 61 / \textbf{57} \\
      \midrule
      Real-robot H2R, EEF (\%)~$\uparrow$ & 11.7 & 18.3 & \textbf{35.0} \\
      GR-1, fine-tuned w/ aux (\%)~$\uparrow$ & -- & 68.9 & \textbf{72.2} \\
      GR-1, fine-tuned w/o aux (\%)~$\uparrow$ & 69.6 & 70.7 & -- \\
      \bottomrule
    \end{tabular*}
  \end{minipage}\hfill
  \begin{minipage}[c]{\dimexpr\textwidth-0.63\textwidth-10pt\relax}
    \includegraphics[width=\linewidth]{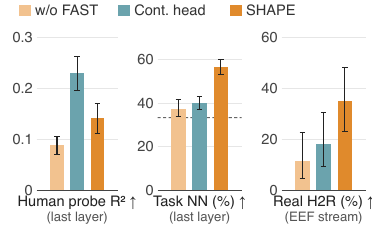}
  \end{minipage}
\end{table}

Dynamically rescaling the L1 weight ($7\text{--}10\times$) to track $\Lfast$'s backbone gradient norm throughout pretraining (Table~\ref{tab:contihead}) separates the two representation properties of Sec.~\ref{sec:repr}: it lifts layer-24 features, matches SHAPE on held-out robot error while lowering human error, and yields higher last-layer linear action probes ($R^2 = 0.230$ vs.\ $0.143$) by emphasizing coarse chunk-mean displacement, yet erodes cross-embodiment task clustering from 66\% at layer 24 down to 40\% at the last layer (matching w/o FAST at 37\%, vs.\ 57\% for SHAPE).

Closed-loop transfer follows task-structure preservation rather than linear probe $R^2$ (Table~\ref{tab:contihead}; Appendix~\ref{supp:contihead}). With zero target-robot demonstrations (H2R, auxiliary action heads disabled for all models), the gradient-matched head reaches 18.3\% vs.\ 35.0\% for SHAPE ($p = 0.055$); on RoboCasa GR-1 (20.8k demos), omitting the head during fine-tuning recovers 70.7\% ($p = 0.11$), whereas retaining it interferes with fine-tuning and drops success to 68.9\% ($p = 0.001$ vs.\ SHAPE's 72.2\%). Conversely, greedily decoding SHAPE's own FAST tokens at test time leaves 15\% of robot windows invalid and trails deterministic zero-noise integration of its EEF stream by 1.4\,mm at the right wrist on valid windows (14.9 vs.\ 14.7\,mm overall against stochastic sampling; Appendix~\ref{supp:fastdec}), confirming that discrete tokens work best as a self-sustaining, structure-preserving backbone target paired with continuous expert execution.

\section{Conclusion}
\label{sec:conclusion}

We presented \textbf{SHAPE}, a dexterous VLA that pairs auxiliary discrete action supervision on the backbone with a shared dual-stream action expert outputting executable EEF and joint streams. In a matched 2$\times$2 study, predicting discrete EEF tokens helps the policy learn from egocentric human data: it drives sustained action gradients on the backbone without dynamic reweighting or conflict with flow matching, and shapes action-informative representations while preserving last-layer task alignment across domains better than continuous regression controls. Sharing the backbone and expert transfers human pretraining to the joint stream without human joint labels, while keeping both streams executable serves complementary deployment pipelines: direct joint readout excels with robot demonstrations, whereas wrist-frame EEF readout via IK enables transfer from human demonstrations alone. Future work will explore scaling this paradigm to larger human video corpora and more robot embodiments.

\setlength{\bibsep}{5pt}
\bibliographystyle{plainnat}
\bibliography{main}

\clearpage
\raggedbottom
\beginappendix
\setcounter{figure}{0}
\setcounter{table}{0}
\renewcommand{\thefigure}{S\arabic{figure}}
\renewcommand{\thetable}{S\arabic{table}}
\renewcommand{\theHfigure}{S\arabic{figure}}
\renewcommand{\theHtable}{S\arabic{table}}

This appendix gives implementation details (Appendix~\ref{supp:impl}), complete real-robot results (Appendix~\ref{supp:real}), the statistical methods (Appendix~\ref{supp:stats}), details of human-to-robot fine-tuning (Appendix~\ref{supp:h2r}), complete offline, representation, and gradient analyses (Appendix~\ref{supp:analyses}), the simulation benchmark (Appendix~\ref{supp:sim}), and the setups of the baselines (Appendix~\ref{supp:baselines}).

\section{Implementation details}
\label{supp:impl}

\subsection{Architecture}
\label{supp:arch}

The backbone is Qwen3-VL-4B~\cite{qwen3vl}: a 0.42B vision encoder and a 4.02B language model with 36 layers and hidden size 2,560. The action expert has the same 36 layers, each attending to the backbone keys and values of its layer, with hidden size 768, feed-forward size 2,752, and 32 query and 8 key--value heads of dimension 128 (0.60B parameters). The flow-matching time enters through adaptive normalization, and inference integrates the flow in 10 Euler steps.

The video head is Wan2.1-T2V-1.3B~\cite{wan} (1.42B parameters with query adapters), with its DiT trained and its VAE frozen. Its 30 DiT blocks cross-attend to the 64 future queries---block $k$ reads backbone layer $k + 6$ after shared normalization and linear projection---and four head-camera motion tokens (replaced by a learned null token when unavailable and in 30\% of samples). Conditioned on the noise-free current latent frame, $\Lvideo$ is the flow-matching loss on the latent encoding of target frames $t + 8$, $t + 16$, $t + 24$, and $t + 32$ at 30\,Hz (arranged as a head-plus-wrists mosaic with wrist tiles masked on human recordings, and noise drawn from $[0.7, 1]$ in 80\% of samples so the DiT relies on the queries).

The sub-task text of a training chunk is the annotated sub-task that overlaps it most (e.g., ``Reach with the right hand, grasp the matte black smartphone from the rear desk, and lift it''); at test time it is decoded greedily.

\subsection{Action and state spaces}
\label{supp:spaces}

Actions are predicted at 60\,Hz in chunks of 64 steps (1.07\,s), and images are sampled at 30\,Hz; at deployment the policy executes 50 steps (0.83\,s) before re-planning. Table~\ref{tab:supp-dims} lists the dimensions of both streams. Two of the seven torso and neck joints never move and are masked. Human recordings carry no torso or neck labels, so these dimensions are masked for human samples.

\begin{table}[H]
  \caption{\textbf{Dimensions of the two action streams.} States use the same layout. Per side, the EEF stream has the wrist position (3), the wrist rotation as the first two rows of its rotation matrix (6), and the positions of the five fingertips in the wrist frame (5 $\times$ 3). The joint stream is not supervised on human pretraining samples.}
  \label{tab:supp-dims}
  \centering\footnotesize
  \begin{tabular*}{\narrowtablewidth}{@{\extracolsep{\fill}}lcc@{}}
    \toprule
    Component & EEF stream & Joint stream \\
    \midrule
    Per side & wrist 3 + 6, tips 5 $\times$ 3 & arm 7, hand 22 \\
    Torso and neck & 7 & 7 \\
    \midrule
    Total & 55 & 65 \\
    Supervised, robot & 53 & 63 \\
    Supervised, human & 48 & 0 \\
    \bottomrule
  \end{tabular*}
\end{table}

\subsection{Normalization}
\label{supp:norm}

In pretraining, each data source is normalized with its own statistics, mapping the 2nd and 98th percentiles of every dimension to $-1$ and 1; human samples use the robot statistics for the dimensions they lack. FAST tokenizes the normalized EEF chunk after clipping it to $[-1, 1]$.

\subsection{Training}
\label{supp:train}

Table~\ref{tab:supp-hparams} lists the optimization hyperparameters for pretraining and downstream fine-tuning.

\begin{table}[H]
  \caption{\textbf{Training hyperparameters.} Fine-tuning lengths are given for the tasks with robot demonstrations and for the H2R tasks. Additional terms in the last row apply chunk temporal smoothing (0.02) and scale human EEF flow matching by 0.5 to temper higher-variance retargeted hand trajectories.}
  \label{tab:supp-hparams}
  \centering\footnotesize
  \begin{tabular*}{\narrowtablewidth}{@{\extracolsep{\fill}}lll@{}}
    \toprule
    & Pretraining & Fine-tuning \\
    \midrule
    Optimizer & \multicolumn{2}{l}{AdamW, $\beta = (0.9, 0.95)$, weight decay $10^{-8}$} \\
    Grad.\ clipping & \multicolumn{2}{l}{global norm 1.0} \\
    Learning rate & $5.7 \times 10^{-5}$ & $5 \times 10^{-5}$ \\
    Expert LR & \multicolumn{2}{l}{0.5 $\times$ learning rate} \\
    Schedule & 5\% warmup, cosine & 5\% warmup, cosine to 1\% \\
    Batch & 1,152 & 128 \\
    Length & 40k steps & 30 epochs; 14,261 steps \\
    EMA decay & \multicolumn{2}{l}{0.9998} \\
    Human\,:\,robot & 1\,:\,1 & one source \\
    $\Lfast$ weight & 0.1 & 0.1; 0 for H2R \\
    $\Lvideo$ weight & \multicolumn{2}{l}{1} \\
    $\Ltext$ weight & \multicolumn{2}{l}{0.2} \\
    Other weights & \multicolumn{2}{l}{smoothing 0.02, human EEF $\times$ 0.5} \\
    \bottomrule
  \end{tabular*}
\end{table}

\subsection{Data}
\label{supp:data}

Table~\ref{tab:models} summarizes the pretraining corpora (73,077 usable human episodes across 320 tasks and 10,505 usable robot episodes across 105 tasks), and Table~\ref{tab:supp-ft} gives the train and validation episode splits for all nine downstream tasks.

\begin{table}[H]
  \caption{\textbf{Fine-tuning data per task.} Entries give the number of episodes.}
  \label{tab:supp-ft}
  \centering\footnotesize
  \begin{tabular*}{\narrowtablewidth}{@{\extracolsep{\fill}}lccc@{}}
    \toprule
    Task & Demonstrations & Train & Validation \\
    \midrule
    Insert Battery & robot & 243 & 6 \\
    Reinsert USB & robot & 202 & 6 \\
    Reinsert Type-C & robot & 221 & 7 \\
    Wipe Bowl & robot & 196 & 5 \\
    Open Cabinet & robot & 447 & 12 \\
    Close Bottle Cap & robot & 413 & 16 \\
    \midrule
    Pick Eggplant & human & 737 & 8 \\
    Pick Cup & human & 1,187 & 29 \\
    Stack Cups & human & 188 & 12 \\
    \bottomrule
  \end{tabular*}
\end{table}

\section{Complete real-robot results}
\label{supp:real}

In Table~\ref{tab:benchmark}, the joint stream outperforms the EEF stream in all model--task pairs except Open Cabinet for w/o FAST (11 vs.\ 13 of 20), and removing $\Lfast$ lowers 22 of the 24 per-task results (except Insert Battery with the EEF stream and human data, tied at 6/20, and Reinsert USB with the joint stream and without human data, 0.25 vs.\ 0.33).

Table~\ref{tab:supp-stages} breaks down stage completion on Reinsert USB and Reinsert Type-C. Without tactile feedback or high-resolution close-up servoing at millimeter tolerance, final connector reinsertion remains a shared bottleneck across all methods (0--3 of 20 trials); SHAPE's score gains come primarily from reliably completing the bimanual unplugging and cable re-grasping stages, where only the joint stream of SHAPE reaches 20/20 on both tasks (compared to 8/20 and 20/20 for $\pi_{0.5}$, 5/20 and 16/20 for GR00T N1.7, 10/20 and 14/20 without human data, and 3/20 and 15/20 without $\Lfast$; without $\Lfast$, only 5--15/20 Type-C trials even complete the initial unplug). Table~\ref{tab:supp-h2r} gives the complete H2R results across both streams of all four models (0/240 for the joint stream vs.\ 32/240 for the EEF stream).

\begin{table}[ht]
  \caption{\textbf{Stage completion on Reinsert USB and Reinsert Type-C.} Entries give trials (out of 20) that complete each stage for the baselines and the four matched models; a trial that completes a stage has completed the earlier ones. A trial earns the credit of the last stage it completes (USB: 0.5 unplug, 1.0 insert; Type-C: 0.3 unplug, 0.6 grasp, 1.0 insert); Table~\ref{tab:benchmark} gives the mean credit.}
  \label{tab:supp-stages}
  \centering\footnotesize
  \setlength{\tabcolsep}{3.2pt}
  \begin{tabular*}{\narrowtablewidth}{@{\extracolsep{\fill}}lcccccc@{}}
    \toprule
    & & \multicolumn{2}{c}{Reinsert USB} & \multicolumn{3}{c}{Reinsert Type-C} \\
    \cmidrule(lr){3-4}\cmidrule(lr){5-7}
    Method & Stream & Unplug & Insert & Unplug & Grasp & Insert \\
    \midrule
    GR00T N1.7 & Joint & 5 & 2 & 16 & 16 & 2 \\
    $\pi_{0.5}$ & Joint & 8 & 2 & 20 & 20 & 1 \\
    \midrule
    SHAPE & EEF & 6 & 0 & 20 & 16 & 0 \\
    & Joint & 20 & 2 & 20 & 20 & 2 \\
    \midrule
    w/o human & EEF & 6 & 0 & 16 & 12 & 0 \\
    & Joint & 10 & 0 & 20 & 14 & 2 \\
    \midrule
    w/o FAST & EEF & 0 & 0 & 6 & 6 & 1 \\
    & Joint & 3 & 0 & 15 & 15 & 3 \\
    \midrule
    w/o both & EEF & 1 & 0 & 5 & 2 & 0 \\
    & Joint & 12 & 1 & 12 & 10 & 1 \\
    \bottomrule
  \end{tabular*}
\end{table}

\begin{table}[ht]
  \caption{\textbf{Human-to-robot transfer across both streams.} Entries give successes out of 20 trials for both readouts of every pretrained model.}
  \label{tab:supp-h2r}
  \centering\footnotesize
  \begin{tabular*}{\narrowtablewidth}{@{\extracolsep{\fill}}lcccc@{}}
    \toprule
    Pretraining & Stream & Eggplant & Cup & Stack \\
    \midrule
    w/o both & EEF & 0 & 0 & 0 \\
    & Joint & 0 & 0 & 0 \\
    \midrule
    w/o human & EEF & 0 & 3 & 1 \\
    & Joint & 0 & 0 & 0 \\
    \midrule
    w/o FAST & EEF & 3 & 2 & 2 \\
    & Joint & 0 & 0 & 0 \\
    \midrule
    SHAPE & EEF & 10 & 8 & 3 \\
    & Joint & 0 & 0 & 0 \\
    \bottomrule
  \end{tabular*}
\end{table}

\section{Statistical methods}
\label{supp:stats}

\paragraph{Score intervals.}
Each policy is run for $n = 20$ trials per task, and each trial earns a score $x \in [0, 1]$: 1 for a success and 0 for a failure, or for Reinsert USB and Reinsert Type-C the credit of the last stage completed (Table~\ref{tab:supp-stages}). If $\mu$ is the mean score of a policy on a task, the variance of $x$ is at most $\mu(1-\mu)$, because $x^2 \leq x$, with equality for binary scores. The six-task score of a policy is $S = \sum_t \mu_t / 6$, where $\mu_t$ is its mean score on task $t$. Since the tasks are run independently, the variance of $S$ is at most $V = \sum_t \mu_t(1-\mu_t)/(36n)$, and we report the 95\% interval $S \pm 1.96\sqrt{V}$. For the total success rates in human-to-robot transfer (Table~\ref{tab:h2r}), we report exact (Clopper--Pearson) 95\% intervals over the 60 trials of the three tasks.

\paragraph{Comparisons.}
With robot demonstrations, two policies with six-task scores $S_1$ and $S_2$ and variance bounds $V_1$ and $V_2$ (or four-task scores restricted to the binary tasks) are compared by a two-sided $z$-test on $z = (S_1 - S_2)/\sqrt{V_1 + V_2}$. In human-to-robot transfer, two policies on one task are compared by a two-sided Fisher exact test on their success counts, and over tasks by the exact conditional Cochran--Mantel--Haenszel (CMH) test stratified by task. Differences and their intervals are computed from unrounded values, so a reported difference can deviate from the difference of the rounded entries in the last digit.

\section{Human-to-robot fine-tuning}
\label{supp:h2r}

\subsection{Setup}
\label{supp:h2r-setup}

Each pretrained model is fine-tuned from its EMA weights on the human recordings of one task only, for 14,261 steps of 128 samples, with $\Lfast$ disabled across all models (Sec.~\ref{sec:h2r}). Images are the head-camera frames of the human capture, cropped to match the field of view of the robot's head camera; because the capture has no wrist cameras, wrist images are masked out of attention both in fine-tuning and on the robot, as for human samples in pretraining. Both streams are supervised: the EEF stream with tracked wrists and retargeted robot fingertips (Sec.~\ref{sec:setup}), and the joint stream with retargeted joints.

\subsection{Why joint actions do not transfer}
\label{supp:joint}

\begin{figure}[t]
  \centering
  \includegraphics[width=0.5\textwidth]{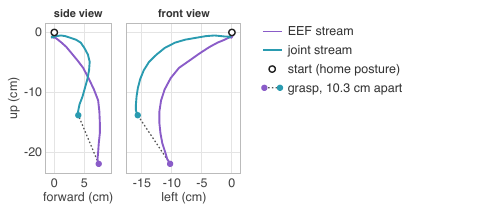}
  \caption{\textbf{Open-loop right-wrist paths from the robot's home posture.} A human Pick Eggplant demonstration is replayed to the grasp in the torso frame. EEF stream: the demonstrated wrist motion applied at the robot's wrist; joint stream: the demonstrated joint offsets added to the home posture.}
  \label{fig:supp-wrist}
\end{figure}

As discussed in Sec.~\ref{sec:stream}, replayed open loop from the robot's home posture without any model, a human demonstration's joint offsets $\Delta q$ place the right hand 10.3\,cm away from where its EEF actions put it at the grasp, on average over the 745 Pick Eggplant demonstrations in Table~\ref{tab:supp-ft} (Fig.~\ref{fig:supp-wrist}). Quantifying the underlying posture gap, retargeted human arm postures lie outside the robot's posture distribution: none is within 10$^\circ$ of a robot posture on all 14 arm joints, whereas 92\% of the robot postures are within 10$^\circ$ of a posture from other robot episodes (median nearest distance 38$^\circ$ vs.\ 6$^\circ$).

\section{Offline, representation, and gradient analyses}
\label{supp:analyses}

\subsection{Offline evaluation}
\label{supp:offline}

For each evaluation window (a fixed starting frame and 64-step target chunk), every model and checkpoint samples one 64-step action chunk with the same random noise, and we compute the mean 3D Euclidean position error of the wrists and fingertips over the 64 steps (converting joint-stream predictions via forward kinematics), paired by task across models. The main test set has six splits and 14,100 windows, read out from the EEF stream: training episodes (808 episodes of 105 tasks, 2,392 windows), new episodes of these tasks (14 tasks, 211 episodes, 2,379 windows), held-out tasks (3 robot tasks absent from pretraining, 814 episodes, 2,395 windows; labeled \textit{held-out tasks (EEF)} in Fig.~\ref{fig:interaction}), robot data of human tasks (5 robot tasks seen only in the human pretraining data, 536 episodes, 2,288 windows), and human recordings from training episodes (236 tasks, 852 episodes, 2,387 windows) and from new episodes (15 tasks, 100 episodes, 2,259 windows).

\begin{figure}[t]
  \centering
  \includegraphics[width=\textwidth]{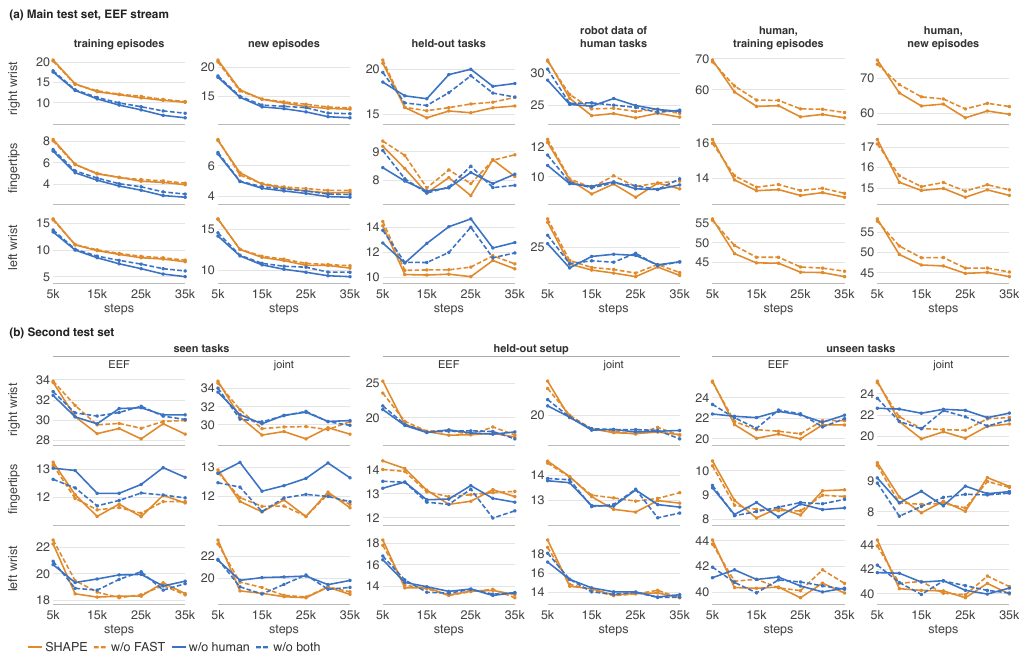}
  \caption{\textbf{Offline action error across pretraining steps.} Errors (mm; lower is better) of the four models are shown for the right wrist, right fingertips, and left wrist (rows). (a) Main test set, EEF stream (splits in Appendix~\ref{supp:offline}); on human recordings, only the two models pretrained on them are shown. (b) Second test set, EEF stream and joint stream through forward kinematics. Orange: with human data; dashed: without FAST.}
  \label{fig:supp-offline}
\end{figure}

The second test set evaluates independent teleoperated robot recordings through both the EEF stream and the joint stream (via forward kinematics) across three splits: unseen validation episodes of 58 pretraining tasks (\textit{seen tasks}; 58 episodes, 577 windows), new-batch and new-scene captures of 4 pretraining task families (\textit{held-out setup}; 120 episodes, 1,088 windows), and 6 tasks absent from all pretraining data (\textit{unseen tasks}; 180 episodes, 1,808 windows, whose joint readout is labeled \textit{unseen tasks (joint)} in Fig.~\ref{fig:interaction}). We report position errors of the right wrist, right fingertips, and left wrist (Fig.~\ref{fig:supp-offline}); CIs are task-level bootstrap.

Across checkpoints from 10k to 35k steps (Fig.~\ref{fig:interaction}a; Fig.~\ref{fig:supp-offline}), pairing $\Lfast$ with human data lowers error relative to w/o FAST at every checkpoint on both human splits (right wrist by 1.5--2.6\,mm, left wrist by 1.1--2.1\,mm, fingertips by 0.2--0.3\,mm; all 95\% CIs $> 0$) as well as on held-out tasks (EEF; $-0.63$\,mm on average), robot data of human tasks ($-0.79$\,mm), and unseen tasks (joint; $-0.62$\,mm). Without human data, $\Lfast$ lowers right-wrist error on seen training and validation episodes ($-0.67$ and $-0.49$\,mm) while raising zero-shot error on held-out tasks (EEF) by +1.10\,mm (Sec.~\ref{sec:offline}). Holding $\Lfast$ fixed (SHAPE vs.\ w/o human), human pretraining reduces right-wrist error by 2.90\,mm on held-out tasks (EEF), 1.05\,mm on robot data of human tasks, and 1.70\,mm on unseen tasks (joint).

\subsection{Representations}
\label{supp:repr}

The probes use the 2,400 windows of Appendix~\ref{supp:offline} (400 per split, 800 of them human). The target is the first eight principal components of the standardized EEF chunk (both wrists and all fingertips, relative to the current state, at 8 of the 64 steps). A probe standardizes the features, keeps 256 principal components, and fits a ridge regression. $R^2$ is cross-validated over five episode folds within a domain; the transfer $r$ fits the probe on robot frames and reports the variance-weighted Pearson correlation on human frames. The shuffled baseline in Fig.~\ref{fig:repr}a permutes the targets within each domain 200 times and plots the largest 95th percentile of $r$ across models (0.041). Image tokens are those of the current and previous head-camera frames; expert features are read at the last denoising step of deployment sampling. CIs are episode bootstrap, paired across models.

The effect of FAST builds up through the language model. On human $R^2$, it is significant from layer 12 on at both checkpoints and largest in the second half of the language model, and the transfer of robot probes rises with it (Table~\ref{tab:supp-probes}); both models without human data stay at $R^2 \leq 0.002$ and $r \leq 0.05$ in every row. At 15k, the prefix mean gives $R^2$ 0.121 vs.\ 0.082 and $r$ 0.273 vs.\ 0.208 for SHAPE and w/o FAST ($\Delta$ +0.039 [+0.029, +0.050] and +0.065 [+0.038, +0.092]). Human data also reach the joint stream of the expert, which they do not supervise in pretraining: on human frames, its features predict the human actions with $R^2$ 0.16 and 0.13 for the two models pretrained with human data, against $\leq 0$ for the two without.

\subsection{Matched tasks}
\label{supp:matched}

Robot windows come from three teleoperated tasks that are absent from robot pretraining but occur in the human pretraining data, which supplies the human windows: sweeping crumbs into a dustpan vs.\ sweeping or mopping the floor, putting plates into a dish rack vs.\ stowing plates and bowls, and moving food from a pan to a plate vs.\ stir-frying and plating. Each task has 240 windows per domain, and every instruction is replaced by ``Do the task'', as the original instructions name the task. With each domain's mean removed, we compute the share of a human window's 10 nearest robot windows (cosine, joint PCA-50) from its task, and the accuracy on human windows of a task classifier fit on robot windows (standardization, PCA-50, logistic regression); CIs are paired bootstrap over human episodes.

\begin{figure}[t]
  \centering
  \includegraphics[width=\textwidth]{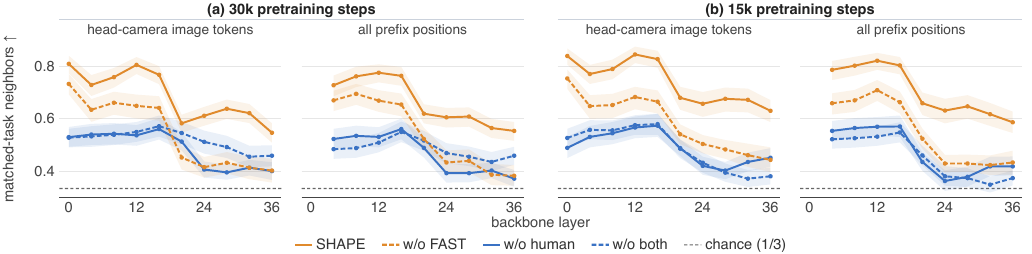}
  \caption{\textbf{Matched-task robot neighbors across layers.} Share of matched-task robot neighbors (Fig.~\ref{fig:repr}b, c) from the vision tower (layer 0) through the language model (layers 4--36), at (a) 30k and (b) 15k pretraining steps across head-camera image tokens and all prefix positions. Shaded bands show 95\% confidence intervals over human episodes.}
  \label{fig:supp-neighbors}
\end{figure}

\begin{table}[ht]
  \caption{\textbf{Linear probes at 30k steps.} Language-model rows use the head-camera image tokens, except the prefix mean of Fig.~\ref{fig:repr}a (last layer, all prefix positions); expert rows use the input projection and the middle and last layers. $\Delta$ = SHAPE $-$ w/o FAST; bold: 95\% CI excludes zero.}
  \label{tab:supp-probes}
  \centering\footnotesize
  \setlength{\tabcolsep}{2.4pt}
  \begin{tabular*}{\narrowtablewidth}{@{\extracolsep{\fill}}lcccccc@{}}
    \toprule
    & \multicolumn{3}{c}{Human $R^2$~$\uparrow$} & \multicolumn{3}{c}{Transfer $r$~$\uparrow$} \\
    \cmidrule(lr){2-4}\cmidrule(lr){5-7}
    Features & \method{} & \makecell[b]{w/o\\FAST} & $\Delta$~$\uparrow$ & \method{} & \makecell[b]{w/o\\FAST} & $\Delta$~$\uparrow$ \\
    \midrule
    Vision tower & 0.028 & 0.020 & \textbf{+0.008} & 0.056 & 0.059 & $-0.003$ \\
    \midrule
    \multicolumn{7}{@{}l}{\textit{Language model}} \\
    Layer 4 & 0.018 & 0.020 & $-0.001$ & 0.058 & 0.051 & +0.007 \\
    Layer 12 & 0.024 & 0.016 & \textbf{+0.008} & 0.055 & 0.038 & +0.017 \\
    Layer 20 & 0.056 & 0.034 & \textbf{+0.022} & 0.101 & 0.064 & \textbf{+0.037} \\
    Layer 28 & 0.143 & 0.092 & \textbf{+0.051} & 0.239 & 0.179 & \textbf{+0.060} \\
    Layer 36 & 0.148 & 0.099 & \textbf{+0.050} & 0.290 & 0.224 & \textbf{+0.066} \\
    Prefix mean & 0.143 & 0.089 & \textbf{+0.054} & 0.281 & 0.205 & \textbf{+0.076} \\
    \midrule
    \multicolumn{7}{@{}l}{\textit{Expert, EEF stream}} \\
    Input & 0.311 & 0.266 & \textbf{+0.045} & 0.463 & 0.420 & \textbf{+0.044} \\
    Middle & 0.297 & 0.253 & \textbf{+0.044} & 0.425 & 0.413 & +0.012 \\
    Last & 0.302 & 0.262 & \textbf{+0.040} & 0.462 & 0.404 & \textbf{+0.058} \\
    \midrule
    \multicolumn{7}{@{}l}{\textit{Expert, joint stream}} \\
    Input & 0.161 & 0.134 & \textbf{+0.027} & 0.272 & 0.255 & +0.017 \\
    Last & 0.160 & 0.133 & \textbf{+0.027} & 0.261 & 0.239 & +0.022 \\
    \bottomrule
  \end{tabular*}
\end{table}

For the head-camera image tokens, SHAPE has the highest matched-task neighbor share at every probed layer of both checkpoints (Fig.~\ref{fig:supp-neighbors}), exceeding w/o FAST with 95\% CIs excluding zero throughout (peaking at +0.21 at layer 32, at both 30k and 15k steps); the robot-trained task classifier agrees from layer 16 on.

\subsection{Gradients}
\label{supp:grad}

For each loss and module, we estimate the expected gradient from 256 batches at 15k and 30k steps and from 128 at the other checkpoints, on human and on robot batches with the same seed, with jackknife standard errors over eight blocks of batches. Each module is represented by one million random coordinates. The energy share of a loss is the squared norm of its weighted expected gradient over the sum for all four losses, with the weights of Eq.~\eqref{eq:loss} and Table~\ref{tab:supp-hparams}. Cosines are between expected gradients and are shown only where both are resolved above batch noise.

$\Lfast$ is the largest source of the language model's gradient at every checkpoint, on human and robot batches alike (47--97\%; Table~\ref{tab:supp-gradshare}), and supplies 17--71\% of the vision tower's. Where resolved, the cosine between the EEF and joint flow-matching gradients is +0.70 to +0.93 on the vision tower and rises from +0.25 at 5k steps to +0.84 at 40k on the expert (Fig.~\ref{fig:grad}c); the two streams agree without FAST too (vision tower +0.80 and +0.79, language model +0.63 and +0.76 at 15k and 30k). Between human and robot batches, the cosine is positive at all six resolved checkpoints on the vision tower (+0.08 to +0.50) and at four of five on the language model (+0.43 to +0.69, and $-0.19 \pm 0.23$ at 35k; Fig.~\ref{fig:grad}d).

\begin{table}[ht]
  \caption{\textbf{Gradient energy share of $\Lfast$ across pretraining.} Entries give the share of the expected gradient energy (\%), after loss weighting, that $\Lfast$ contributes to the language model and the vision tower of SHAPE, on human (H) and robot (R) batches. The action expert receives only $\mathcal{L}_{\text{FM}}$.}
  \label{tab:supp-gradshare}
  \centering\footnotesize
  \begin{tabular*}{\narrowtablewidth}{@{\extracolsep{\fill}}lcccc@{}}
    \toprule
    & \multicolumn{2}{c}{Language model} & \multicolumn{2}{c}{Vision tower} \\
    \cmidrule(lr){2-3}\cmidrule(lr){4-5}
    Steps & H & R & H & R \\
    \midrule
    5k & 79 & 52 & 21 & 17 \\
    10k & 68 & 52 & 31 & 19 \\
    15k & 58 & 61 & 26 & 50 \\
    20k & 54 & 47 & 30 & 67 \\
    25k & 89 & 97 & 46 & 42 \\
    30k & 70 & 93 & 36 & 45 \\
    35k & 78 & 84 & 31 & 62 \\
    40k & 82 & 78 & 71 & 53 \\
    \bottomrule
  \end{tabular*}
\end{table}

\subsection{Decoding the FAST tokens}
\label{supp:fastdec}

As a deployment ablation testing why SHAPE executes the continuous expert rather than decoding its auxiliary tokens (Sec.~\ref{sec:deploy}), we greedily decode the FAST head on the 2,400 windows of Appendix~\ref{supp:offline} (400 per subset) at 30k and 40k steps over the 1,024 FAST ids and EOS (up to 512 tokens) and detokenize valid sequences into $64 \times 55$ EEF chunks (returning the midpoint of the normalization range when invalid). On the same backbone pass, we compare against the EEF stream of the expert, both sampled with random noise and integrated deterministically from zero initial noise (\textit{zero-noise expert}).

\begin{table}[ht]
  \caption{\textbf{Greedy decoding of the FAST tokens.} Evaluation on the windows of Appendix~\ref{supp:offline}. Decoded: share of windows whose tokens form a valid chunk; TF acc.: next-token accuracy given ground-truth prefixes; $\Delta$: right-wrist error of SHAPE's decoded tokens minus that of its zero-noise expert, on decoded robot windows (mm; 95\% CI over episodes).}
  \label{tab:supp-fastdec}
  \centering\footnotesize
  \setlength{\tabcolsep}{2.5pt}
  \begin{tabular*}{\narrowtablewidth}{@{\extracolsep{\fill}}lcccc@{}}
    \toprule
    Pretraining & Steps & \makecell[b]{Decoded~$\uparrow$\\robot / human} & \makecell[b]{TF acc.~$\uparrow$\\robot / human} & $\Delta$~$\downarrow$ \\
    \midrule
    SHAPE & 40k & 0.85 / 0.84 & 0.45 / 0.28 & +1.4 [+1.0, +1.8] \\
    & 30k & 0.81 / 0.74 & 0.45 / 0.27 & +1.5 [+1.0, +1.9] \\
    \midrule
    w/o human & 40k & 0.82 / 0.00 & 0.46 / 0.06 & -- \\
    & 30k & 0.84 / 0.03 & 0.45 / 0.07 & -- \\
    \bottomrule
  \end{tabular*}
\end{table}

Autoregressive token decoding is less reliable and less accurate than the continuous expert (Table~\ref{tab:supp-fastdec}): at 40k steps, 15\% of robot windows fail to form a valid chunk, overall robot right-wrist error is 14.9\,mm vs.\ 14.7\,mm (sampled) and 12.4\,mm (zero-noise), and even on valid windows decoded tokens trail the zero-noise expert by 1.4\,mm; 30k shows the same pattern. Without human pretraining data, the FAST head decodes only 1 of 800 human windows (next-token accuracy 0.06 vs.\ 0.28 for SHAPE), whereas on robot data the two models decode alike. Discrete FAST tokens are therefore best used as a training-time backbone target while leaving test-time execution to the continuous expert.

\subsection{Continuous-head controls on the backbone}
\label{supp:contihead}

\paragraph{Control design.}
To separate the role of backbone gradient magnitude from that of discrete autoregressive prediction, we train two matched pretraining controls that replace the FAST tokens with 64 learnable queries in the exact same positions of the backbone sequence, mutually visible to one another and masked from the action expert as in SHAPE, and regress the normalized $64 \times 55$ EEF chunk from their last-layer hidden states via a two-layer MLP under an L1 loss (network initialization, random seeds, and data order match SHAPE and w/o FAST). The \textit{fixed-weight} control sets $\lambda = 0.1$ (trained to 15k steps) so that its unweighted backbone gradient norm matches $\Lfast$ at initialization. Because L1 regression saturates early, however, its backbone gradient norm shrinks to $0.11\text{--}0.13\times$ that of $\Lfast$ from 5k steps onward: at 15k steps it supplies only 2\% and 8\% of the language model's expected gradient energy on human and robot batches (vs.\ 58\% and 61\% for $\Lfast$) and 1\% of the vision tower's (vs.\ 26\% and 50\%). It recovers the last-layer prefix-mean probe ($R^2 = 0.117$ vs.\ $0.121$ for SHAPE and $0.082$ for w/o FAST) and matched-task neighbor share ($0.603$ vs.\ $0.608$), but leaves intermediate probes at the w/o FAST level (layer-24 head-camera tokens: $0.042$ vs.\ $0.046$ for w/o FAST and $0.083$ for SHAPE) and recovers only half of $\Lfast$'s open-loop error drop on human training episodes ($55.63$\,mm vs.\ $54.72$\,mm for SHAPE and $56.72$\,mm for w/o FAST).

\paragraph{Pretraining error and representations.}
The \textit{gradient-matched} control (trained for the full 40k steps) therefore rescales $\lambda(t) \in [0.1, 1.0]$ every 50 steps using an exponential moving average of the unweighted L1 backbone gradient norm (RMS across human and robot batches) to track the weighted backbone gradient norm of $\Lfast$ in SHAPE; across all eight checkpoints from 5k to 40k steps, the realized ratio is $0.92\text{--}1.03\times$ ($\lambda(t)$ peaks near $0.97$ over 1.5k--7.5k steps and settles near $0.73$ from 20k to 40k steps). Restoring the gradient budget lifts intermediate probes (at 15k, layer-24 human $R^2$ rises from $0.042$ to $0.101$, and human training-episode error drops to $52.66$\,mm). At 30k steps (Table~\ref{tab:supp-contihead}), direct L1 coordinate regression yields lower open-loop right-wrist error than SHAPE on in-distribution robot and human splits, while staying within 1\,mm of SHAPE on held-out robot tasks and on the second test set from 15k steps onward, where SHAPE keeps a small edge on the joint readout of unseen tasks at 30k steps (0.50\,mm). In representations, both models lift layer-24 probes and matched-task neighbor share well above w/o FAST, but diverge at the last layer: the gradient-matched head nearly doubles human $R^2$ on the chunk-mean displacement component, raising overall prefix-mean human $R^2$ ($0.230$ vs.\ $0.143$) and transfer $r$ ($0.437$ vs.\ $0.281$), whereas last-layer prefix-mean matched-task neighbors drop from $0.603$ (fixed weight) to $0.457$ at 15k and $0.398$ at 30k (matching w/o FAST at $0.375$, vs.\ $0.566$ for SHAPE) and robot-to-human task-classifier accuracy falls below w/o FAST ($0.382$ vs.\ $0.459$, vs.\ $0.535$ for SHAPE).

\begin{table}[t]
  \caption{\textbf{Gradient-matched continuous-head control vs.\ SHAPE and w/o FAST.} Pretraining metrics are evaluated at 30k steps (Appendices~\ref{supp:offline}--\ref{supp:grad}); downstream closed-loop evaluations start from the pretrained EMA checkpoints, following Appendix~\ref{supp:h2r-setup} for real-robot H2R transfer (60 trials across 3 tasks; auxiliary action heads disabled for all models), Table~\ref{tab:benchmark} for three real-robot tasks with robot demonstrations (60 trials across Insert Battery, Reinsert USB, and Wipe Bowl; 30-epoch fine-tuning retaining each model's pretraining auxiliary action head), and Appendix~\ref{supp:sim} for RoboCasa GR-1 (mean of 80k, 90k, and 100k checkpoints over 1,200 task--seed pairs). $\Delta$ denotes SHAPE minus Cont.\ head (negative favors SHAPE on error rows $\downarrow$, positive favors SHAPE on $\uparrow$ rows; bold marks $\Delta$ and the better of the two values where the 95\% CI excludes zero and $p < 0.05$). On H2R, the three-task score CI excludes zero while the stratified exact CMH test gives $p = 0.055$; $^\dagger$compares SHAPE fine-tuned w/ $\Lfast$ against Cont.\ head fine-tuned w/o aux.}
  \label{tab:supp-contihead}
  \centering\footnotesize
  \setlength{\tabcolsep}{2.5pt}
  \begin{tabular*}{\textwidth}{@{\extracolsep{\fill}}lccccc@{}}
    \toprule
    Metric & SHAPE & w/o FAST & Cont.\ head & $\Delta$ & 95\% CI ($p$) \\
    \midrule
    \multicolumn{6}{@{}l}{\textit{Backbone gradient budget and open-loop right-wrist error at 30k steps (mm~$\downarrow$)}} \\
    Weighted VLM grad.\ norm / $\Lfast$ & $1.00\times$ & -- & $0.98\times$ (fixed $0.13\times$) & -- & -- \\
    Training episodes (EEF) & 10.62 & 10.84 & \textbf{9.47} & \textbf{+1.15} & [+0.92, +1.41] \\
    Human, new episodes (EEF) & 60.67 & 62.88 & \textbf{59.19} & \textbf{+1.48} & [+0.37, +2.63] \\
    Held-out tasks (EEF) & 15.70 & 16.33 & 16.10 & $-0.40$ & [$-2.40$, +1.93] \\
    Robot data of human tasks (EEF) & 23.72 & 24.37 & 23.44 & +0.28 & [$-0.22$, +0.81] \\
    Second test set, unseen tasks (EEF) & 21.35 & 21.76 & 21.23 & +0.12 & [$-1.97$, +1.02] \\
    Second test set, unseen tasks (joint) & \textbf{20.95} & 21.65 & 21.45 & {\boldmath$-$}\textbf{0.50} & [$-1.66$, $-0.002$] \\
    \midrule
    \multicolumn{6}{@{}l}{\textit{Backbone and expert representations at 30k steps ($\uparrow$)}} \\
    Human $R^2$, layer 24 (head-cam) & \textbf{0.107} & 0.052 & 0.091 & \textbf{+0.015} & [+0.002, +0.028] \\
    Human $R^2$, prefix mean & 0.143 & 0.089 & \textbf{0.230} & {\boldmath$-$}\textbf{0.087} & [$-0.105$, $-0.069$] \\
    Transfer $r$, layer 24 (head-cam) & 0.167 & 0.099 & 0.172 & $-0.005$ & [$-0.030$, +0.022] \\
    Transfer $r$, prefix mean & 0.281 & 0.205 & \textbf{0.437} & {\boldmath$-$}\textbf{0.156} & [$-0.182$, $-0.129$] \\
    Human $R^2$, expert EEF stream (last) & 0.302 & 0.262 & \textbf{0.324} & {\boldmath$-$}\textbf{0.021} & [$-0.036$, $-0.006$] \\
    Matched-task NN, layer 24 (head-cam) & 0.612 & 0.415 & \textbf{0.663} & {\boldmath$-$}\textbf{0.052} & [$-0.085$, $-0.019$] \\
    Matched-task NN, prefix mean & \textbf{0.566} & 0.375 & 0.398 & \textbf{+0.168} & [+0.138, +0.198] \\
    Robot$\to$human task acc., prefix mean & \textbf{0.535} & 0.459 & 0.382 & \textbf{+0.153} & [+0.086, +0.220] \\
    \midrule
    \multicolumn{6}{@{}l}{\textit{Closed-loop downstream transfer ($\uparrow$; 3-task score for real-robot demos, \% otherwise)}} \\
    Real-robot H2R, EEF (3 tasks) & 35.0 & 11.7 & 18.3 & +16.7 & [+1.8, +31.6] ($p = 0.055$) \\
    Robot demos, Joint, w/ aux (3 tasks) & \textbf{0.75} & 0.36 & 0.35 & \textbf{+0.40} & [+0.24, +0.56] ($p < 0.001$) \\
    GR-1, fine-tuned w/ aux: All & \textbf{72.2} & -- & 68.9 & \textbf{+3.3} & [+1.3, +5.3] ($p = 0.001$) \\
    \quad Close / PnP & 66.6 / \textbf{74.1} & -- & 64.9 / 70.3 & +1.7 / \textbf{+3.8} & -- \\
    GR-1, fine-tuned w/o aux: All & -- & 69.6 & 70.7 & +1.6$^\dagger$ & [$-0.4$, +3.5] ($p = 0.11$) \\
    \quad Close / PnP & -- & 57.1 / 73.8 & 63.7 / 73.0 & +2.9 / +1.1$^\dagger$ & -- \\
    \bottomrule
  \end{tabular*}
\end{table}

\paragraph{Closed-loop transfer.}
Starting from the 40k-step EMA checkpoint of the gradient-matched control, we evaluate closed-loop transfer across real-robot H2R transfer (Appendix~\ref{supp:h2r-setup}), real-robot manipulation with robot demonstrations (Table~\ref{tab:benchmark}), and the 24 RoboCasa GR-1 tasks (Appendix~\ref{supp:sim}; Table~\ref{tab:supp-contihead}). On the three real-robot H2R tasks---where fine-tuning uses identical supervision with auxiliary action heads disabled across all models, isolating pretraining representations with zero target-robot demonstrations---the gradient-matched control succeeds in 5/20 trials on Pick Eggplant, 5/20 on Pick Cup, and 1/20 on Stack Cups (11/60 = 18.3\%, vs.\ 21/60 = 35.0\% for SHAPE, stratified exact CMH $p = 0.055$, three-task score $\Delta = +16.7$ points, 95\% CI [+1.8, +31.6]; and vs.\ 7/60 = 11.7\% for w/o FAST, CMH $p = 0.44$). On three real-robot tasks with robot demonstrations (Insert Battery, Reinsert USB, and Wipe Bowl; 20 joint-stream trials each after full 30-epoch fine-tuning retaining the continuous L1 head at its end-of-pretraining weight $0.735$, mirroring SHAPE's retention of $\Lfast$ in Table~\ref{tab:benchmark}), the gradient-matched control achieves 5/20 on Insert Battery, a mean stage score of 0.35 on Reinsert USB (12/20 unplug, 2/20 insert), and 9/20 on Wipe Bowl, for a three-task mean score of 0.35---matching w/o FAST (0.36: 11/20, 0.08, 9/20; $p = 0.92$) and trailing SHAPE by 0.40 (0.75: 16/20, 0.55, 18/20; 95\% CI [+0.24, +0.56], $p < 0.001$ by the test of Appendix~\ref{supp:stats}). On RoboCasa GR-1 (20.8k target-robot demonstrations), we evaluate both a matched \textit{fine-tuned w/ aux} arm---retaining the continuous L1 head on normalized forward-kinematics EEF chunks (51 active GR-1 dimensions) at weight $0.735$---and a \textit{fine-tuned w/o aux} arm that omits it as in w/o FAST. Consistent with the real-robot benchmark result, retaining the continuous head reaches 68.9\% overall, falling 3.3 points short of SHAPE ($72.2\%$, $p = 0.001$) and matching w/o FAST ($69.6\%$, $\Delta = -0.7$ points, $p = 0.52$). Omitting the head during fine-tuning reaches 70.7\%, statistically indistinguishable from both w/o FAST ($+1.0$ points, $p = 0.32$) and SHAPE ($-1.6$ points, $p = 0.11$); keeping the continuous head in fine-tuning raises validation video loss by 6--7\% and lowers PnP by 2.7 points ($70.3\%$ vs.\ $73.0\%$, $p = 0.026$). Both continuous-head arms recover most of SHAPE's advantage over w/o FAST on the six two-stage articulated Close tasks ($64.9\%$ and $63.7\%$ vs.\ $57.1\%$, $p \le 0.002$) and exceed w/o human ($65.2\%$, $p < 0.001$) and $\pi_{0.5}$ ($64.8\%$, $p < 0.001$). Together, these controls show that sustained direct action gradients on the backbone account for much of the intermediate-layer encoding and high-demonstration transfer gains, while discrete token prediction realizes them without external gradient-norm tracking, preserves last-layer matched-task structure, and remains compatible with downstream fine-tuning.

\section{Simulation benchmark}
\label{supp:sim}

\begin{table}[t]
  \caption{\textbf{Per-task success rates (\%) on the 24 RoboCasa GR-1 tabletop tasks.} Entries average 50 evaluation seeds across the three checkpoints of Appendix~\ref{supp:sim} for both external baselines and the four matched models.}
  \label{tab:supp-sim-tasks}
  \centering\footnotesize
  \begin{tabular*}{\textwidth}{@{\extracolsep{\fill}}llcccccc@{}}
    \toprule
    Group & Task & GR00T N1.7 & $\pi_{0.5}$ & w/o both & w/o human & w/o FAST & SHAPE \\
    \midrule
    \multirow{6}{*}{Close} & Bottle $\to$ Cabinet Close & 80.7 & 76.7 & 74.0 & 75.3 & 72.7 & \textbf{82.7} \\
    & Can $\to$ Drawer Close & \textbf{86.0} & 44.7 & 76.0 & 73.3 & 60.0 & 78.0 \\
    & Cup $\to$ Drawer Close & \textbf{51.3} & 26.7 & 48.0 & 49.3 & 42.0 & 48.7 \\
    & Milk $\to$ Microwave Close & \textbf{71.3} & 50.7 & 56.7 & 60.0 & 62.7 & 66.0 \\
    & Potato $\to$ Microwave Close & 70.7 & 56.7 & 66.0 & 70.7 & 60.7 & \textbf{74.0} \\
    & Wine $\to$ Cabinet Close & \textbf{55.3} & 53.3 & 48.0 & 46.7 & 44.7 & 50.0 \\
    \midrule
    \multirow{18}{*}{PnP} & Cuttingboard $\to$ Basket & 82.7 & 82.0 & 78.7 & 77.3 & \textbf{86.7} & 82.7 \\
    & Cuttingboard $\to$ Cardboardbox & 49.3 & \textbf{80.7} & 52.7 & 70.7 & 74.7 & 74.0 \\
    & Cuttingboard $\to$ Pan & 92.7 & \textbf{95.3} & 90.0 & 92.0 & 89.3 & 92.0 \\
    & Cuttingboard $\to$ Pot & 71.3 & \textbf{75.3} & 61.3 & 62.0 & 71.3 & 74.0 \\
    & Cuttingboard $\to$ Tieredbasket & 41.3 & 54.0 & 55.3 & 51.3 & 59.3 & \textbf{64.7} \\
    & Placemat $\to$ Basket & 79.3 & \textbf{83.3} & 67.3 & 59.3 & 72.7 & 78.0 \\
    & Placemat $\to$ Bowl & 41.3 & 68.0 & 59.3 & 62.7 & \textbf{72.7} & 72.0 \\
    & Placemat $\to$ Plate & 48.0 & 76.0 & 57.3 & 72.7 & \textbf{82.0} & 74.7 \\
    & Placemat $\to$ Tieredshelf & 27.3 & 32.7 & 34.7 & 32.0 & \textbf{49.3} & \textbf{49.3} \\
    & Plate $\to$ Bowl & 56.0 & 75.3 & 73.3 & 83.3 & \textbf{88.7} & 88.0 \\
    & Plate $\to$ Cardboardbox & 44.7 & 67.3 & 58.0 & 53.3 & 67.3 & \textbf{71.3} \\
    & Plate $\to$ Pan & 39.3 & 76.7 & 60.7 & 66.0 & \textbf{83.3} & 80.0 \\
    & Plate $\to$ Plate & \textbf{89.3} & 72.0 & 72.7 & 80.7 & 80.0 & 82.0 \\
    & Tray $\to$ Cardboardbox & 73.3 & 70.7 & 70.7 & 67.3 & \textbf{80.0} & \textbf{80.0} \\
    & Tray $\to$ Plate & 87.3 & 72.0 & 75.3 & 85.3 & 84.0 & \textbf{90.7} \\
    & Tray $\to$ Pot & 73.3 & 70.0 & 63.3 & 69.3 & 74.7 & \textbf{76.0} \\
    & Tray $\to$ Tieredbasket & 64.0 & 59.3 & 60.7 & \textbf{64.7} & 64.0 & 60.0 \\
    & Tray $\to$ Tieredshelf & 26.7 & 35.3 & 28.0 & 39.3 & \textbf{48.7} & 44.7 \\
    \midrule
    \multicolumn{2}{@{}l}{Close mean (6 tasks)} & \textbf{69.2} & 51.4 & 61.4 & 62.6 & 57.1 & 66.6 \\
    \multicolumn{2}{@{}l}{PnP mean (18 tasks)} & 60.4 & 69.2 & 62.2 & 66.1 & 73.8 & \textbf{74.1} \\
    \multicolumn{2}{@{}l}{All mean (24 tasks)} & 62.6 & 64.8 & 62.0 & 65.2 & 69.6 & \textbf{72.2} \\
    \bottomrule
  \end{tabular*}
\end{table}

\paragraph{Tasks.}
The GR-1 tabletop tasks of RoboCasa~\cite{robocasa,gr00tn1} control a \mbox{GR-1} humanoid with two six-DoF hands from its head camera. Six articulated tasks place an object into a cabinet, drawer, or microwave and close it (Close in Table~\ref{tab:sim}); the other 18 are pick-and-place tasks that move an object from one receptacle to another (PnP). We fine-tune all models on the GR00T-Teleop-Sim collection~\cite{gr00tn1}, which has 20,802 training and 480 validation episodes across the 24 tasks, with identical train/validation splits and normalization statistics.

\paragraph{Fine-tuning.}
All four matched models use the exact same fine-tuning recipe, starting from their pretrained EMA checkpoints. The joint stream predicts the 29 joints of the GR-1 (7 per arm, 6 per hand, 3 for the waist) through its pretrained output layer, with the other dimensions masked, in chunks of 64 steps, of which 12 are executed; the policy receives no proprioceptive state. Models pretrained with $\Lfast$ keep it on EEF chunks computed by forward kinematics (wrist pose and five fingertips per hand, in the head-camera frame), and $\Lvideo$ predicts head-camera frames up to 32 steps ahead for all four models. We train for 100k steps with batch 256 and a base learning rate of $5 \times 10^{-5}$ (cosine decay after a 5\% warmup), using $0.25\times$ the base learning rate for the action expert with its first decoder layer frozen, and $10\times$ for the joint output layer. Both external baselines ($\pi_{0.5}$ and GR00T N1.7) are fine-tuned on the exact same data split and 29-DoF relative joint targets under the same 100k-step, batch-256 budget (Appendix~\ref{supp:baselines}).

\paragraph{Evaluation.}
Every model is evaluated with EMA weights on 50 episodes per task (seeds 0--49, at most 720 simulation steps, executing 12 steps per inference); Table~\ref{tab:sim} and Table~\ref{tab:supp-sim-tasks} report the mean across the 80k, 90k, and 100k checkpoints. An episode seed fixes the initial scene, so any two models are compared by a paired $t$-test over the 1,200 task--seed pairs, each averaged over the three checkpoints. The 95\% intervals of the overall success rates (Table~\ref{tab:sim}) are normal intervals of the mean over the same pairs. Across task groups (Table~\ref{tab:supp-sim-tasks}), with 20.8k demonstrations ($\sim$870 per task), single-stage pick-and-place (PnP) is largely saturated once human pretraining is included (74.1\% for SHAPE vs.\ 73.8\% for w/o FAST, both well above 66.1\% for w/o human), whereas the two-stage articulated Close tasks still benefit substantially from $\Lfast$ (+9.5 points, 66.6\% vs.\ 57.1\%).

\section{Baselines}
\label{supp:baselines}

Both baselines start from their public checkpoints and are fine-tuned in full; Table~\ref{tab:supp-baselines} lists their setups. On the six real-robot tasks, they train on the demonstrations of Table~\ref{tab:supp-ft} for 30 epochs, as SHAPE does, using the same three cameras and predicting the same 65 relative joint actions as our joint stream (with the two immobile torso/neck joints masked), and are evaluated with their final checkpoints; each baseline retains its native pretrained chunk length and official fine-tuning learning-rate schedule (Table~\ref{tab:supp-baselines}) for stable convergence. To keep the pretrained 32-d action projections of $\pi_{0.5}$, each control step is split into a left-limb and a right-limb token as in Ego-Pi~\cite{egopi}, so a chunk of 25 steps becomes 50 tokens. On the simulated GR-1 benchmark (Appendix~\ref{supp:sim}), both baselines train for 100k steps at batch 256 on the single head camera and the 29 active GR-1 relative joint targets (with unused slots masked) alongside their native state inputs (discrete state prompt for $\pi_{0.5}$, state encoder with 0.2 dropout for GR00T N1.7), and are evaluated under the exact same EMA protocol as our models (executing 12 steps per inference).
\begin{table}[ht]
  \caption{\textbf{Setups of the baselines on the six tasks with robot demonstrations.} Human data: hours of egocentric human video in pretraining. Chunks are in steps at 60\,Hz, and steps range over the tasks.}
  \label{tab:supp-baselines}
  \centering\footnotesize
  \begin{tabular*}{\narrowtablewidth}{@{\extracolsep{\fill}}lcc@{}}
    \toprule
    & $\pi_{0.5}$ & GR00T N1.7 \\
    \midrule
    Initialization & \multicolumn{2}{c}{public checkpoint} \\
    Human data & none & 20k\,h \\
    Cameras & \multicolumn{2}{c}{head and both wrists} \\
    Actions & \multicolumn{2}{c}{65 joints, relative to the state} \\
    Chunk & 25 & 40 \\
    \midrule
    Batch & 256 & 256 \\
    Steps & 20k--72k & 20k--72k \\
    Learning rate & $2.5 \times 10^{-5}$ & $10^{-4}$ \\
    Warmup & 1k steps & 2k steps \\
    Cosine decay to & 10\% & 20\% \\
    \bottomrule
  \end{tabular*}
\end{table}

\end{document}